\documentclass[10pt,twocolumn,letterpaper]{article}

\usepackage{cvpr}              %
\usepackage{soul}
\usepackage{xcolor}
\usepackage{verbatim}
\usepackage{etoolbox}
\usepackage{colortbl}
\usepackage{multirow}
\usepackage{siunitx}
\usepackage{etoolbox}
\makeatletter
\patchcmd{\@maketitle}{\vspace*{24pt}}{\vspace*{6pt}}{}{}
\makeatother

\usepackage{soul}
\usepackage{xcolor}
\usepackage{etoolbox}
\usepackage{colortbl}
\usepackage{multirow}
\usepackage{siunitx}
\usepackage{etoolbox}
\usepackage{xspace}
\usepackage{float}
\usepackage{caption}
\usepackage{footmisc}

\newcommand{\fref}[1]{Fig. \ref{#1}}

\newcommand{\vggts}{VGGT${}^\star$\xspace}
\newcommand{\method}{Co-Me\xspace}
\definecolor{ourcolor}{HTML}{94b6d2}
\definecolor{r1clr}{HTML}{E63C41}
\definecolor{r2clr}{HTML}{3CB44B}
\definecolor{r3clr}{HTML}{4169E1}

\newcommand{\compactpar}[1]{\vspace{2mm}\noindent\textbf{#1}\;\;}

\makeatletter
\renewcommand{\abstract}{%
   \iftoggle{cvprpagenumbers}{}{%
     \thispagestyle{empty}
   }
   \centerline{\large\bf Abstract}%
   \vspace{6pt}
   \noindent%
   \it\ignorespaces%
}
\makeatother

\definecolor{cvprblue}{rgb}{0.21,0.49,0.74}
\usepackage[pagebackref,breaklinks,colorlinks,allcolors=cvprblue]{hyperref}

\def\paperID{14918} %
\def\confName{CVPR}
\def\confYear{2025}

\title{
\method: Confidence Guided Token Merging for Visual Geometric Transformers\\
\vspace*{-4pt}
\iftoggle{cvprfinal}{
    {\normalsize\href{https://co-me-tokens.github.io}{co-me-tokens.github.io}}
}{
    {\normalsize\texttt{[URL REDACTED DUE TO ANONYMOUS POLICY]}}
}
}

\author{
Yutian Chen${}^{1, 2}$ \\
\texttt{\small yutianch@andrew.cmu.edu}
\and
Yuheng Qiu${}^{1}$\\
\texttt{\small yuhengq@andrew.cmu.edu}
\and
Ruogu Li${}^{1}$\\
\texttt{\small ruoguli@andrew.cmu.edu}
\vspace{0.01cm}
\and
Ali Agha${}^2$\\
\texttt{\small ali@fieldai.com}
\and
Shayegan Omidshafiei$^2$\\
\texttt{\small shayegan@fieldai.com}
\and
Jay Patrikar${}^{2}$\\
\texttt{\small jay@fieldai.com}
\and
Sebastian Scherer${}^{1, 2}$\\
\texttt{\small basti@fieldai.com}\\
}

\begin{document}

\twocolumn[{
\maketitle
\iftoggle{cvprfinal}{
\vspace{-1cm}
\begin{center}
\large
\noindent$^1$Carnegie Mellon University \hspace{5em} 
$^2$Field AI 
\end{center}
\vspace{-0.5cm}
}{}

\begin{center}
    \includegraphics[width=\linewidth]{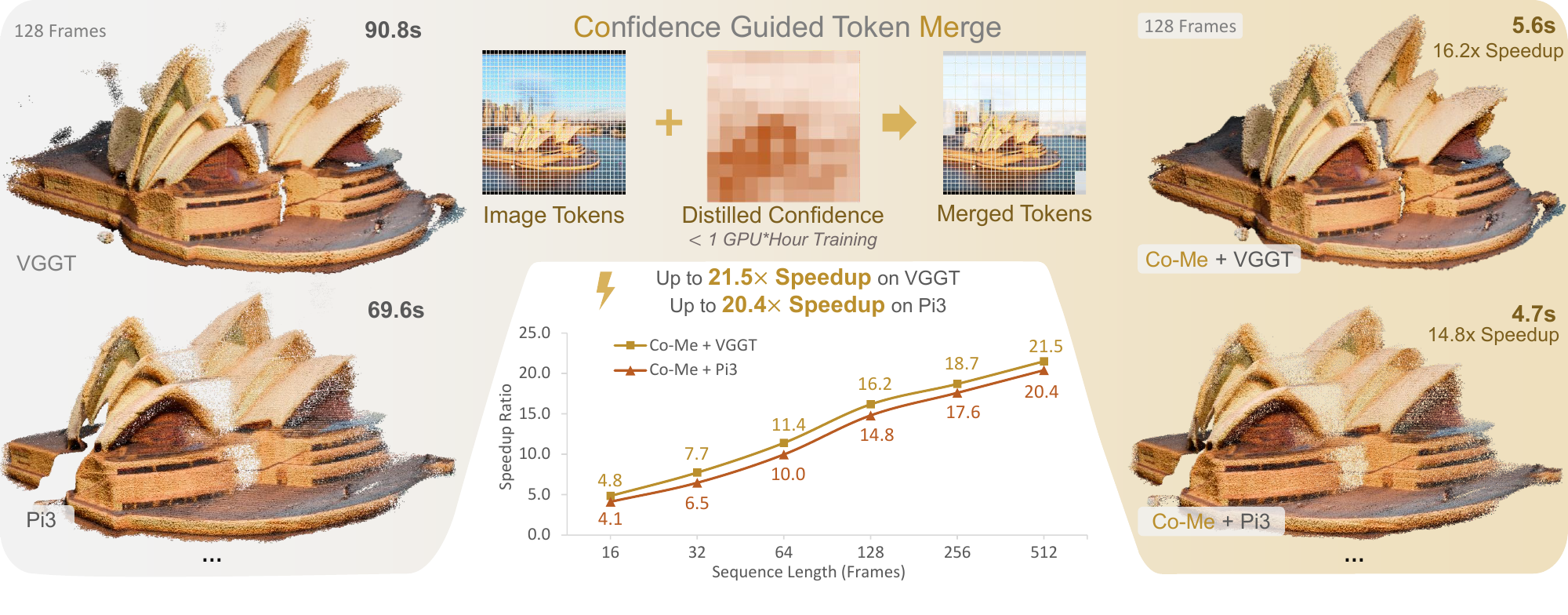}
\end{center}
\captionsetup{type=figure}
\vspace{-1cm}
\captionof{figure}{%
    \textbf{\normalsize\method} accelerates visual geometric transformers by selectively merging low-confidence tokens guided by a distilled confidence predictor.
    When applied to VGGT and Pi3, \method achieves up to $21.5\times$ and $20.4\times$ speedup without retraining or architectural changes to the ViT backbone, turning geometric transformers into real-time-capable models for 3D perception.
}\label{fig:teaser}
\vspace{0.25cm}
}]

\begin{abstract}
We propose \underline{Co}nfidence-Guided Token \underline{Me}rging (\method), an acceleration mechanism for visual geometric transformers without retraining or finetuning the base model. 
\method distilled a light-weight confidence predictor to rank tokens by uncertainty and selectively merge low-confidence ones, effectively reducing computation while maintaining spatial coverage.
Compared to similarity-based merging or pruning, the confidence signal in \method reliably indicates regions emphasized by the transformer, enabling substantial acceleration without degrading performance.
\method applies seamlessly to various multi-view and streaming visual geometric transformers, achieving speedups that scale with sequence length. 
When applied to VGGT and Pi3, \method achieves up to $21.5\times$ and $20.4\times$ speedup, making visual geometric transformers practical for real-time 3D perception and reconstruction.
\end{abstract}

\vspace{-0.5cm}
\vspace{-0.1cm}
\section{Introduction}
\vspace{-0.1cm}
\label{sect: 1-introduction}
Reasoning about 3D structure from visual input serves as a fundamental capability for intelligent systems, such as autonomous navigation, robotic manipulation, and augmented reality.
Recent breakthroughs in visual geometry models, exemplified by VGGT~\cite{wang2024vggt}, $\pi^3$~\cite{wang2026pi3permutationequivariantvisualgeometry}, MapAnything ~\cite{keetha2025mapanythinguniversalfeedforwardmetric}, and DepthAnything 3~\cite{depthanything3}, have shown remarkable progress in scene reconstruction and understanding tasks.
However, these advances come at a significant computational cost:  Vision Transformers (ViTs) incur quadratic complexity with respect to the input sequence length. This severely limits real-time deployment in resource-constrained environments.
To empower embodied intelligence with fast and accurate 3D reconstruction, there is a critical need for acceleration methods that preserve geometric understanding while reducing computational cost.

The main barrier of efficient ViT inference lies in the quadratic time complexity $O(n^2d)$ of attention with respect to the number of tokens $n$ and feature dimension $d$~\cite{vaswani2023attentionneed}. Although efficient attention mechanisms like FlashAttention~\citep{dao2023flashattention2fasterattentionbetter} reduce memory overhead, their computational complexity remains high. 
Other than the attention, the multi-layer perceptron (MLP) also takes a considerable amount of computation~\citep{9522921}.
This motivates token pruning and merging, which directly reduces the number of tokens to mitigate quadratic cost while maintaining similar performance.

A major line of work, exemplified by DynamicViT~\cite{rao2021dynamicvit}, progressively removes uninformative tokens to accelerate ViT inference. 
However, such approaches are mainly effective for inherently sparse tasks like image classification~\cite{rao2021dynamicvit, yu2023xprunerexplainablepruningvision, Yin_2022_CVPR} or segmentation~\cite{tang2023dynamictokenpruningplain}, as discarding tokens in dense geometric tasks often eliminate contextual cues required by accurate 3D reconstruction.
Additionally, these methods need costly retraining, which is impractical as model and dataset scales grow, e.g., foundation model like VGGT already approach a billion parameters~\cite{wang2024vggt}.

Compared to token pruning, token merging offers a more balanced acceleration strategy; however, existing practices mostly rely on heuristics. 
For instance, ToMe~\cite{bolya2023tome} merges tokens according to feature similarity for image classification, while FastVGGT~\cite{shen2025fastvggt} leverages both feature norms and cosine similarity to guide the merging process in the global attention operator.
These methods are effective in scenarios with extremely long inputs, such as a 1,000-image sequence. 
However, the speedups in reality are modest since global attention only constitutes a fraction of runtime when using efficient attention with moderate-length sequences.

Inspired by human foveal vision~\cite{10.1167/jov.20.12.2}, where high-acuity processing targets key regions while peripheral areas are coarsely perceived, we aim to reduce computation without sacrificing 3D reconstruction fidelity.
A key observation is that most image tokens in ViTs do not actively contribute to 3D reconstruction.
We further find that the high-confidence region predicted by the network strongly correlates with the region that the ViT emphasizes.
In contrast, low-confidence regions often correspond to background, which provides coarse contextual cues rather than precise geometric estimation.
Moreover, these regions exhibit poor quality and are typically discarded by downstream tasks like 3D reconstruction~\cite{deng2025vggtlongchunkitloop, wang2024vggt,keetha2025mapanythinguniversalfeedforwardmetric} or visual SLAM~\citep{qiu2025mac}.
These findings raise a fundamental question: 
\textbf{How can we identify and reduce redundant tokens in visual geometry transformers without compromising geometric fidelity?} %

In this work, we propose a novel \textit{confidence-guided token merging} approach for accelerating visual geometric transformers.
We observed that the confidence jointly estimated from the visual geometric model suggests the necessary information for scene understanding.
Based on this observation, we distilled a confidence module that predicts per-patch confidence rankings in self-supervised manner.
Guided by the distilled confidence, our method selectively merges low-confidence tokens, reducing computation in both the attention and MLP without sacrificing reconstruction quality.
Compared to existing methods, our method preserves high-fidelity results in geometrically critical areas while significantly reducing the inference time.
This efficiency further enables practical on-device deployment: when integrated with MapAnything, our accelerated model runs on an \texttt{\small NVIDIA Jetson Thor} at 3.5FPS with chunked 4-frame input and is $1.5\times$ faster than the original model, showing its suitability for real-time deployment.

The main contributions of this work are:
\begin{itemize}

\item We propose a novel confidence-guided token merging method that selectively merges tokens in low-confidence regions, delivering significant acceleration without retraining or architectural changes to the foundation model. 

\item We introduce a self-supervised confidence distillation module that estimates per-patch confidence rankings from intermediate encoder features to guide token merging.

\item Experiments show that our method produces consistent speedup across various visual geometry transformers and input conditions with minimal performance degradation. 

\item We implement a CUDA kernel and TensorRT~\citep{tensorrt} plugin to minimize the runtime overhead. We have further deployed and validated our method on the edge device.

\end{itemize}

{\color{ourcolor}
\textbf{This is an extended version of the conference paper.} We extended or modified the paper in following aspects:}

\begin{itemize}
    \item We identified and fixed a bug in the benchmarking code for MapAnything and update the results in \cref{sect:experiments}.
\end{itemize}

\section{Related Works}

\subsection{Visual Geometric Transformer and Confidence}

Visual geometric transformers have revolutionized geometric understanding by enabling single-pass 3D reconstruction without iterative optimization.
DUSt3R~\cite{wang2024dust3r} first demonstrated the pairwise 3D point-map regression, while MASt3R~\cite{wang2024master} incorporated explicit confidence modeling to improve geometric reliability.
Recent extensions, including MUSt3R~\cite{cabon2025must3rmultiviewnetworkstereo}, Spann3R~\cite{wang2024spann3r}, CUT3R~\cite{wang2025continuous}, and more~\citep{jang2025pow3rempoweringunconstrained3d,lu2024align3ralignedmonoculardepth,dong2025reloc3rlargescaletrainingrelative,lan2025stream3rscalablesequential3d,yang2025fast3r3dreconstruction1000} further generalize this paradigm to multi-view and streaming settings. Built upon these developments, VGGT~\cite{wang2024vggt} and more ~\cite{keetha2025mapanythinguniversalfeedforwardmetric, wang2026pi3permutationequivariantvisualgeometry, depthanything3} unify camera pose, intrinsics, depth, and point-map prediction within a 1B-parameter ViT, thereby culminating the feed-forward paradigm and achieving state-of-the-art 3D reconstruction.

Crucially, these visual geometric transformers inherently predict confidence maps that quantify the reliability of their predictions~\cite{wang2024vggt, keetha2025mapanythinguniversalfeedforwardmetric, depthanything3, wang2026pi3permutationequivariantvisualgeometry}. High-confidence regions typically correspond to well-textured and stable areas where multi-view cues are consistent, while low-confidence regions emerge in occluded, textureless, or ill-posed zones such as sky or reflective surfaces~\cite{kendall2017uncertaintiesneedbayesiandeep,poggi2020uncertaintyselfsupervisedmonoculardepth,qiu2025mac}.
Despite this rich confidence information, current models allocate uniform computation to all tokens, leading to inefficiency where equal resources are allocated to both geometrically reliable and uncertain regions.
Such uniform processing paradigm offers an opportunity for acceleration, given the quadratic complexity of ViT with respect to token count.

\begin{figure*}
    \centering
    \includegraphics[width=\linewidth]{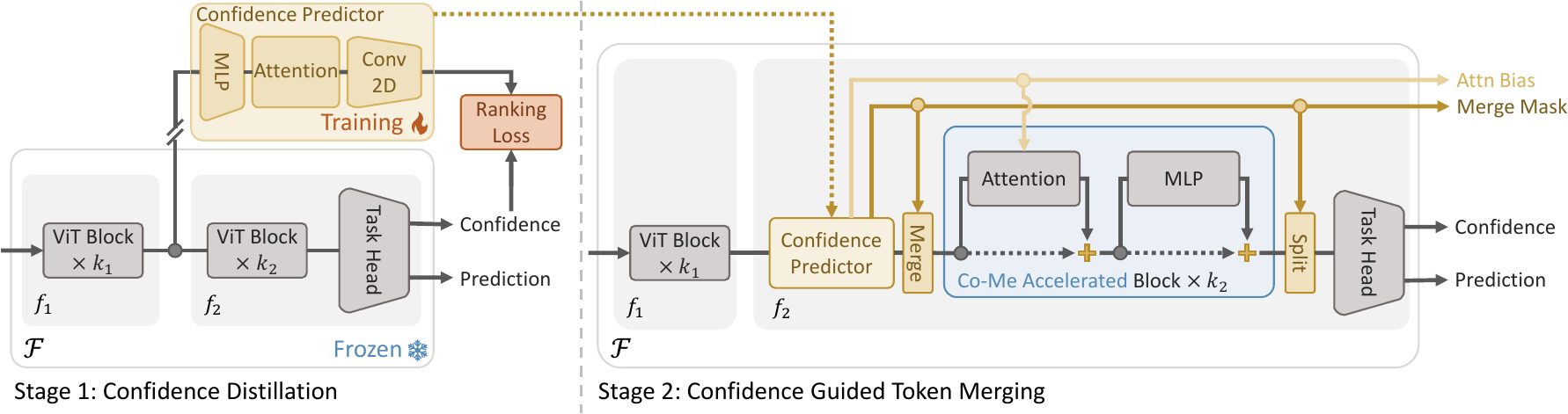}
    \vspace{-16pt}
    \caption{Overview of \method. A lightweight module distilled from the frozen ViT backbone predicts per-token confidence from intermediate features. The predicted confidence is converted into a binary mask that guides token merging on the attention and MLP modules. 
    }
    \label{fig:method-overview}
    \vspace{-6pt}
\end{figure*}

\subsection{Token Pruning and Merging}

Recognizing that token importance varies across different regions, many studies accelerate ViT inference by processing tokens non-uniformly, primarily in 2D vision tasks such as image classification and segmentation. These approaches adapt the number of active tokens and generally fall into two categories: token pruning and token merging.

Token pruning methods dynamically drop tokens during inference to improve computation efficiency. Progressive pruning methods such as SparseViT~\citep{Chen_sparsevit2023}, DynamicViT~\citep{rao2021dynamicvit}, and A-ViT~\citep{Yin_AViT2022} estimate token importance and selectively remove less significant tokens during forward passes, achieving substantial speedups in image classification.
However, Liu et al.~\citep{liu2024revisiting} revealed the fundamental limitation of token pruning on dense prediction tasks such as instance segmentation.
Their study shows consistent performance degradation, which could only be partially alleviated by token reactivation strategies.
Token pruning fundamentally suffers from spatial information loss, critical for maintaining spatial resolution and per-pixel consistency required by dense tasks like 3D reconstruction.

Token merging offers a promising alternative that aggregates similar tokens instead of discarding them.
While this strategy preserves spatial coverage, it achieves a lower acceleration ratio due to the retained tokens.
ToMe~\citep{bolya2023tome} pioneered training-free acceleration for off-the-shelf ViTs by merging tokens based on feature similarity, while advanced techniques like TokenLearner~\citep{ryoo2021tokenlearner}, PuMer~\citep{cao2023pumer}, and ToFu~\citep{kim2023tofu} further refined the merging process for various applications.
FastVGGT~\citep{shen2025fastvggt} introduces similarity and norm-based token merging in the global attention of visual geometric transformers, but its acceleration remains limited since it requires an extremely long input sequence (1,000 frames) to yield a notable speedup.
Our approach builds upon token merging while overcoming its limitations through two key innovations: a distilled confidence module that predicts per-token confidence to guide processing, and a confidence-guided merging strategy that preserves precision for 3D reconstruction with significantly less compute.
This confidence-guided paradigm represents a novel intersection of geometric understanding and computational efficiency, accelerating visual geometric transformers without finetuning the base ViT while addressing the unique demands of real-time 3D reasoning for embodied intelligence.

\vspace{-4pt}
\section{Method}

Illustrated in \fref{fig:method-overview}, \method contains two stages. In the first stage, we distill a light-weight confidence prediction module from the original ViT model (\cref{ssect:confidence-distill}). In the second stage, we use the predicted per-patch confidence score to generate a merge mask during inference, and use this mask to guide the token merging and splitting (\cref{ssect:confidence-guided-token-merging}). Additionally, we employed several efficient implementations to minimize the overhead of token merging (\cref{ssect:efficient-impl}).

\subsection{Confidence Distillation}
\label{ssect:confidence-distill}
To avoid the dilemma of needing result of full inference to accelerate inference, token merging must rely on confidence estimates available \emph{beforehand}. 
Since the encoder features already contain rich cues of confidence estimation, we distill a model that predicts per-token confidence from these features. This enables confidence-guided merging in the remaining part of the network.
Formally, given a network $\mathcal{F}$ that predicts a confidence map $\mathcal{C}$ with input $x$, 
we split it into two parts $\mathcal{F}\!=\!f_2 \circ f_1$. 
The goal is to distill a lightweight network $f'\!\!: \mathrm{Im}\;f_1 \mapsto \mathcal{\mathcal C'}$ that estimates per patch confidence map $\mathcal{C}'$ 
, which resembles $\mathcal C$ on the token-level. Importantly, we never update or back-propagate through the visual geometric model $\mathcal{F}$; all training is confined to the predictor $f'$.

\begin{figure*}
\centering
    \centering
    \includegraphics[width=\linewidth]{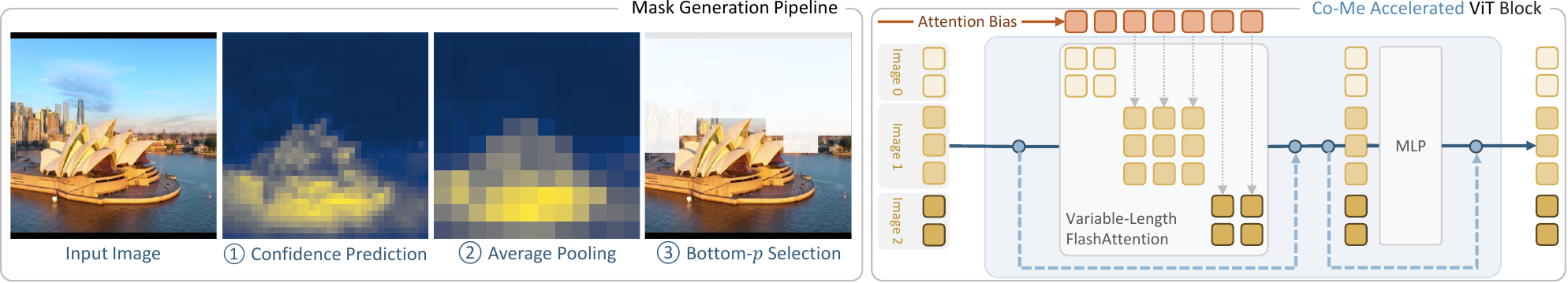}
    \vspace{-16pt}
    \caption{
    The proposed mask generation (left) and accelerated inference block (right).
    Each image sequence generate an individual mask via confidence ranking and bottom-$p$ selection, with per-image token counts varies within the sequence. 
    To inference efficiently over non-uniform batch, We implemented a FlashAttention variant that supports attention bias correction with variable sequence lengths.
    }
    \label{fig:merge-split-operator}
    \vspace{-6pt}
\end{figure*}

\compactpar{Model Design} 
The confidence predictor consists of three lightweight components.
First, an MLP layer projects encoder features into a compact latent space.
Next, a single-head attention captures interactions between patches across frames, enabling global reasoning at low computational cost.
Finally, a Conv2D head compresses the tokens into a confidence map while suppressing spatial noise and promoting smoother predictions.
Such a distilled module only adds less than $2\%$ runtime compared to the full network.\footnote{Measured on VGGT under input sequence length of 128.}

Since the confidence predictor has limited modeling capacity, we relax supervision so that it focuses on learning relative confidence ordering between tokens. 
This relaxation is valid because \method only requires knowledge of the relative ordering to identify which tokens exhibit lower confidence than others.
Specifically, we employed a logistic ranking loss~\cite{burges2005learningtorank} instead of the direct mean square error (MSE), and we define the loss function $\mathcal{L}(\mathcal{C}', \mathcal{C})$ as:
\begin{equation}
\begin{aligned}
\mathcal{L}(\mathcal{C}', \mathcal{C}) = \frac{1}{|\mathcal{P}|}
\sum_{(i,j)\in\mathcal{P}}{
\log\!\left(1 + \exp(\mathcal{C}_{j}' - \mathcal{C}_{i}'))\right)},\\
\mathcal{P} \sim \mathrm{UniformSubset}(\{(i,j) \mid \overline{\mathcal{C}_{i}} > \overline{\mathcal{C}_{j}}\}).
\end{aligned}
\end{equation}
Where $\mathrm{Avg}(\cdot)$ represents the average pooling that takes the average over all pixels in each patch as the patch's overall confidence score. We further verify the advantage of this relaxed loss function over direct MSE loss in \cref{appendix:confidence-distill-loss-ablate}.

\compactpar{Training Details} 
Since the distillation aims to replicate the confidence of the original ViT model, the training is entirely self-supervised and does not rely on any ground-truth labels. 
For the experiments in this paper, we used the TartanAir~\citep{wang2020tartanairdatasetpushlimits} dataset, a synthetic dataset containing more than 500,000 sequential images from diverse environments and motion patterns, for confidence distillation. The distillation converges within approximately 2,000 steps and takes less than an hour on a single \texttt{\small NVIDIA H100 80G HBM3} GPU\footnote{We generate the input-confidence dataset beforehand and do not count inference of acceleration target (e.g. VGGT) in the training time.}.

For all visual geometric transformers, we insert the confidence predictor at the middle of the encoder as this provides a good trade-off between confidence prediction quality and acceleration. We provide an ablation on layer selection in \cref{appendix:additional_analysis_insert_layer}. In all experiments in this paper, the predictor generalizes to unseen data without finetuning.

\subsection{Confidence-Guided Token Merging}
\label{ssect:confidence-guided-token-merging}

The key insight is that merging low-confidence tokens barely affects the predicted geometry in high-confidence regions. 
Since we can receive a confidence prediction only after inference on $f_1$, all operations described in this section are only applied to the remaining part $f_2$ of the network.

\compactpar{Mask Generation}  
Illustrated in~\cref{fig:merge-split-operator}, we construct a binary merging mask under a predefined parameter merge ratio~$p$ from predicted per-token confidence scores.
We first partition tokens along the spatial order into fixed-size groups of $n$ image tokens.
For each group, if the average confidence falls below the $p$-th percentile across all groups in the image sequence, it is marked for merging.
This design allows variable merge ratio across images in the sequence, aligning with the real world use case where certain images in a sequence may contain more information than others.

\compactpar{Token Merging}
Before inference on $f_2$, we apply token-merging to reduce the number of tokens and thereby accelerate inference. 
In the merge operator, each group of image tokens is either aggregated by taking average of all tokens in the group or preserved based on the merge flag.
Formally, for a group of $n$ tokens $G_i$, if the merge flag $m_i$ is \textit{true}, we replace the group with their average; otherwise, $G_i$ remains unchanged. This operation is applied to all groups, and the results are concatenated into a contiguous tensor:
\begin{equation}
\begin{aligned}
    &\mathrm{MergeGrp}(G_i, m_i) = \begin{cases}
        \left\{\frac{1}{n}\sum_{x \in G_i} x\right\} &\text{if }m_i\\
        G_i                        &\text{otherwise}
    \end{cases}\\
    &\mathrm{Merge}(\{G_i\}, \{m_i\}) = \mathrm{Cat}\!\left(\{\mathrm{MergeGrp}(G_i, m_i)\}\right)
\end{aligned}
\end{equation}

\compactpar{Token Splitting}
After inference on $f_2$, the processed token sequence is restored to its original shape through a splitting step. If a processed token group $G'_i$ was not merged in the previous step, we copy it to its original position; otherwise, we replicate the merged token $G_i'=\{x\}$ for $n$ times and place them at their original index.
\begin{equation}
\begin{aligned}
    &\mathrm{SplitGrp}(G'_i, m_i) = \begin{cases}
        \{x, \dots, x\} &\text{if }m_i\\
        G'_i             &\text{otherwise}
    \end{cases}\\
    &\mathrm{Split}(\{G'_i\}, \{m_i\}) = \mathrm{Cat}\!\left(\{\mathrm{SplitGrp}(G'_i, m_i)\}\right)
\end{aligned}
\end{equation}
Such replication-based splitting, inspired by ToMeSD~\citep{Bolya_2023}, allows \method to reduce computation cost while maintaining compatibility with downstream prediction heads.

\begin{figure}
\centering
    \centering
    \includegraphics[width=\linewidth]{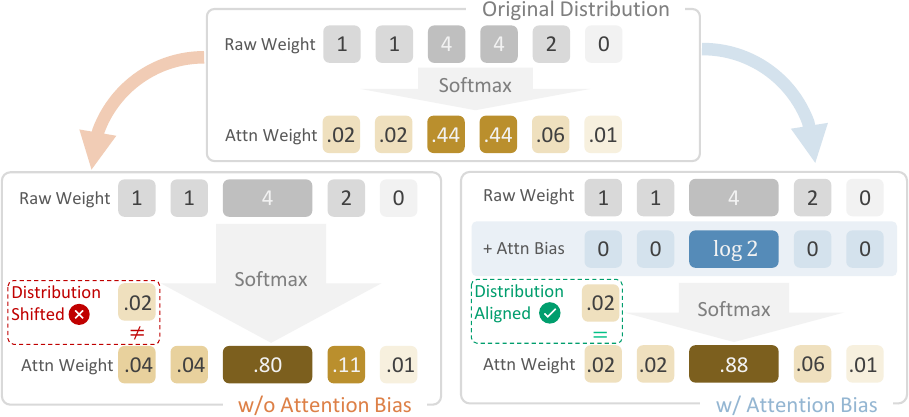}
    \vspace{-12pt}
    \caption{
        Effect of attention bias correction. Merging tokens distorts the weight distribution after softmax operator (left). Adding bias term $\log n$ aligns the merged distribution with original (right).
    }
    \label{fig:attention-bias-correction}
    \vspace{-6pt}
\end{figure}

\compactpar{Attention Bias Correction}
As shown in \fref{fig:attention-bias-correction}, merging tokens into one concentrates multiple attention weights into a single entry. This causes the softmax operator to suppress that entry's normalized attention weight and distorts the distribution.
To counteract this, we introduce an \textit{attention bias correction} that compensates for the merged entries. Specifically, for a merged group with $n$ tokens and raw attention logit $a_i$, we add a bias term $\log n$ to get the corrected $\tilde{a}_i = a_i + \log n.$ Since softmax is exponential, adding $\log n$ to the merged logit scales its weights by $n$ and effectively restores the same total mass that $n$ individual logits contributed before:
\begin{equation}
    \mathrm{softmax}(\tilde{a}_i)
    = \frac{e^{a_i + \log n}}{\sum_j e^{a_j}}
    \approx \sum_{k\in G_i} \frac{e^{a_k}}{\sum_j e^{a_j}}.
\end{equation}
This correction realigns the post-softmax attention weights distribution with the original distribution. 
Although the attention bias introduces additional memory access and slows down the attention operator, our ablation in \cref{sect:analysis}, H4. demonstrates that it significantly improves the performance.

\subsection{Efficient Implementation}
\label{ssect:efficient-impl}
To minimize runtime overhead, we adopt several engineering optimizations. First, we create a variable-length FlashAttention~\cite{dong2024flexattentionprogrammingmodel} kernel that supports per-key attention bias correction.
Second, we exploit this deterministic index relation between original and merged token indices to perform merge and split without the expensive \texttt{Cat} operator.
This index compute is further accelerated with a single-pass exclusive scan~\citep{Merrill2016SinglepassPP}. %
Lastly, we create TensorRT plugin to minimize the latency on resource-limited edge platforms.

\section{Experiments}
\label{sect:experiments}

In this section, we show that our method accelerates both the VGGT~\citep{wang2024vggt}, $\pi^3$~\citep{wang2026pi3permutationequivariantvisualgeometry}, MapAnything (MA)~\citep{keetha2025mapanythinguniversalfeedforwardmetric}, and DepthAnything 3 (DA3)~\citep{depthanything3} with minimal performance change across three downstream tasks on NYUd-v2~\citep{Silberman:ECCV12}, ETH3D~\citep{schoeps2017cvpr}, DTU-MVS~\citep{aanaes2016large}, KITTI Depth~\citep{Uhrig2017THREEDV}, and RealEstate-10K~\citep{zhou2018re10k}. 
These datasets span diverse domains, covering both indoor and outdoor scenes with varying motion, depth range, and length, thus providing a comprehensive evaluation of our method.

\subsection{Experiment Setup}
\label{ssect:experiment-setup}

\compactpar{Baselines} 
We evaluate \method by applying it to VGGT, $\pi^3$, MapAnything, and DepthAnything 3 to measure speedup and performance on depth, camera pose, and point cloud prediction tasks. 
For a fair comparison, we strengthen the VGGT baseline by replacing its naive attention with FlashAttention and incorporating the VRAM optimization trick from FastVGGT, yielding an enhanced version denoted as \vggts. 
Other methods already use an efficient fused attention, so we didn't modifiy their implementation. For simplicity, we evaluate all baselines with image-only input despite some has the ability to receive additional modality as pre-conditioning, potentially improves the output.

Both our method, FastVGGT, and ToMeSD~\citep{Bolya_2023} use a hyperparameter in the range $(0,1)$ to control the merge ratio. We set the merge ratio to $0.5$ for \method and ToMeSD, and $0.9$ for FastVGGT, following the default value in its official implementation.
This setup is more favorable to the FastVGGT baseline, as a higher merge ratio allows more aggressive token reduction.
We fix the group size $n$ to $3\times3$ for all experiments in this paper unless otherwise stated.

\compactpar{Environment} All experiments are conducted on \texttt{\small NVIDIA H100 80GB HBM3} GPU and \texttt{\small Intel Xeon 8468} CPU.

\subsection{Depth Estimation}
\label{ssect:depth-estimation}

\begin{table*}
    \centering
    \resizebox{0.9\linewidth}{!}{%
        \begin{tabular}{l|cccc|cccc|cccc|cccc}
        \toprule
        \multirow{2}{*}{Method} & \multicolumn{4}{c|}{NYUd-v2 (1 Frame)} & \multicolumn{4}{c|}{ETH3D (1 Frame)} & \multicolumn{4}{c|}{DTU-MVS (32 Frames)} & \multicolumn{4}{c}{KITTI Depth (48 Frames)}\\
        & Latency & Speedup & L1$\downarrow$ & $\delta_{1.25}\uparrow$ & Latency & Speedup & L1$\downarrow$ & $\delta_{1.25}\uparrow$ & Latency & Speedup & L1$\downarrow^\star$ & $\delta_{1.25}\uparrow$ & Latency & Speedup & L1$\downarrow$ & $\delta_{1.25}\uparrow$\\
        \midrule
        VGGT            & 66.7  & 1.00$\times$  & 0.186 & 0.940 & 68.4  & 1.00$\times$  & 0.254 & 0.959 & 6094  & 1.00$\times$  & 0.89  & 0.990 & 13045 & 1.00$\times$  & 4.647 & 0.562 \\
        \vggts          & 56.1  & 1.18$\times$  & 0.186 & 0.940 & 59.8  & 1.14$\times$  & 0.254 & 0.959 & 1266  & 4.81$\times$  & 0.89  & 0.990 & 2281  & 5.72$\times$  & 4.647 & 0.562 \\
        ToMeSD$_{0.5}$  & 139   & 0.48$\times$  & 0.221 & 0.925 & 140   & 0.48$\times$  & 0.238 & 0.959 & 1091  & 5.59$\times$  & 0.93  & 0.990 & 1813  & 7.19$\times$  & 4.660 & 0.550 \\
        Fast VGGT       & -     & -             & -     & -     & -     & -             & -     & -     & 2003  & 3.04$\times$  & 0.94  & 0.990 & 3416  & 3.82$\times$  & 4.611 & 0.562 \\
        \rowcolor{ourcolor!30}
        \textbf{Ours}   & 61.5  & 1.09$\times$  & 0.225 & 0.918 & 60.3  & 1.13$\times$  & 0.261 & 0.964 & 788   & 7.73$\times$  & 1.10  & 0.987 & 1313  & 9.94$\times$  & 4.727 & 0.558 \\
        \midrule
        $\pi^3$         & 56.9  & 1.00$\times$  & 0.171 & 0.948 & 58.1  & 1.00$\times$  & 0.257 & 0.953 & 4648  & 1.00$\times$  & 1.63  & 0.987 & 10139 & 1.00$\times$  & 4.684 & 0.589 \\
        \rowcolor{ourcolor!30}
        \textbf{Ours}   & 42.2  & 1.09$\times$  & 0.221 & 0.928 & 49.0  & 1.19$\times$  & 0.270 & 0.948 & 718   & 6.47$\times$  & 2.09  & 0.983 & 1162  & 8.72$\times$  & 4.966 & 0.559 \\
        \midrule
        DA3-Giant       & 52.8  & 1.00$\times$  & 0.272 & 0.899 & 53.1  & 1.00$\times$  & 0.314 & 0.956 & 1576  & 1.00$\times$  & 1.40  & 0.986 & 2461  & 1.00$\times$  & 4.405 & 0.607 \\
        \rowcolor{ourcolor!30}
        \textbf{Ours}   & 51.9  & 1.02$\times$  & 0.463 & 0.792 & 51.3  & 1.04$\times$  & 0.350 & 0.928 & 692   & 2.28$\times$  & 1.78  & 0.984 & 1112  & 2.21$\times$  & 4.453 & 0.590 \\
        \midrule
        MA              & 56.5 & 1.00$\times$  & 0.428 & 0.898 & 54.8 & 1.00$\times$  & 0.421 & 0.939 & 3470 & 1.00$\times$  & 4.59  & 0.965 & 7225 & 1.00$\times$  & 8.867 & 0.280 \\
        \rowcolor{ourcolor!30}
        \textbf{Ours}   & 50.2 & 1.13$\times$  & 0.417 & 0.889 & 49.9 & 1.10$\times$  & 0.445 & 0.929 & 730 & 4.75$\times$  & 6.80  & 0.884 & 1165 & 6.20$\times$  & 10.01 & 0.252 \\
        \bottomrule
        \multicolumn{8}{l}{\small $^\star$DTU-MVS shows L1 depth error in unit of \si{\centi\meter} to preserve sufficient significant digits.}
        \end{tabular}%
    }
    \vspace{-7pt}
    \captionsetup{font=footnotesize}
    \caption{
        Evaluation of Latency (\si{\milli\second}), L1 depth error (\si{\meter}), and $\delta_{1.25}$ (unitless) of \textbf{depth estimation} with scale alignment over NYUd-v2, ETH3D, DTU-MVS, and KITTI Depth datasets. FastVGGT does not support inferring single image.
    }
    \label{tab:depth-evaluation}
    \vspace{-8pt}
\end{table*}

Depth estimation aims to predict a dense per-pixel depth map from one or multiple input images. 
We evaluate our method on both monocular and multi-view settings, following same input setup as the original models, respectively.

\compactpar{Metrics} 
We employed L1 and $\delta_{1.25}$ as depth estimation metrics to capture the absolute and relative depth accuracy. 
Global scale alignment is applied to resolve the scale ambiguity. 
For all methods, we only evaluate regions where tokens are not merged by our method to ensure a fair comparison.
Specifically, given the ground truth depth $d$ and predicted depth $\hat d$, the metrics are defined as:
\begin{equation}
\begin{aligned}
    &L_1=\frac{1}{N}\sum_{i=1}^N \bigl|\hat{d}_i - d_i\bigr|,\\
    &\delta_{1.25}=\frac{1}{N}\sum_{i=1}^N\mathbf{1}\!\left[ \max\!\left(\dfrac{\hat{d}_i}{d_i},\dfrac{d_i}{\hat{d}_i}\right)\!<\!1.25 \right],
\end{aligned}
\end{equation}
where $\mathbf{1}[\cdot]$ is the indicator function.

\compactpar{Performance}
\cref{tab:depth-evaluation} shows the \textit{monocular} depth estimation result on NYUd-v2 and ETH3D and multi-view depth estimation result on DTU-MVS and KITTI Depth. Our method achieves up to up to $1.13\times$ speedup even in single-frame input scenario, where ToMeSD and FastVGGT are either slower or simply fails. In multi-view depth estimation task, our method continuously deliver the best speedup, significantly surpass other baseline methods while maintaining comparable performance in L1 and $\delta_{1.25}$ metrics.

\compactpar{Analysis} Speedup and accuracy retention vary across datasets. In KITTI, image tokens have less spatial overlap, so token merging causes larger information loss. In contrast, datasets like NYUd-v2 and DTU have more redundant fields of view, allowing our method to be more effective.

\subsection{Pose Estimation}
\label{ssect:pose-estimation}

The pose estimation task seeks to predict the 6-DoF camera positions and orientations for all input views, providing a consistent frame for 3D reconstruction and understanding.

\compactpar{Metrics} Following VGGT~\cite{wang2024vggt}, we employed area under curve for relative translation accuracy at 10\si{\centi\meter} (AUC$^t_{10}$) and relative rotary accuracy at 10\si{\degree} (AUC$^r_{10}$) as pose evaluation metrics. $\mathrm{Sim}(3)$ Umeyama alignment~\citep{Lawrence_2019} is applied to remove scale and reference frame ambiguity. Specifically, the metrics are defined as follows:
\begin{equation}
\begin{aligned}
    &\mathrm{AUC}^t_{10} = \int_{0}^{10}{\frac{1}{|\mathcal P|}\sum_{i, j\in \mathcal P}{\mathbf{1}\left[\|t_{i,j} - \hat{t}_{i, j}\|_2 < x\right]} \;\mathbf{d}x} \\
    &\mathrm{AUC}^r_{10} = \int_{0}^{10}{\frac{1}{|\mathcal P|}\sum_{i, j\in \mathcal P}{\mathbf{1}\left[\angle(R_{i, j}^{-1} \; \hat{R}_{i, j}) < x\right]}\; \mathbf{d}x},
\end{aligned}    
\end{equation}
where $\hat{t}_{i, j}$ and $t_{i, j}$ denote the predicted and ground-truth relative translation between frame $i, j$ and $\hat{R}_{i, j}, R_{i, j}$ the corresponding rotations. $\angle(\cdot)$ measures the geodesic angle between rotations and $\mathcal{P}$ includes all unordered camera pairs.

\compactpar{Performance} \cref{tab:pose-evaluation} reports the performance on DTU and RealEstate-10K\footnote{We only evaluate on the first 100 samples due to runtime limitation.} (RE10K). 
Compared to VGGT, our method achieves $7.72\times$ and $16.2\times$ speedups on DTU (32 frames) and RE10K (128 frames) with minimal drop in AUC$^t_{10}$ and AUC$^r_{10}$. 
Comparing to FastVGGT, our method attains higher acceleration with comparable pose accuracy. 

\compactpar{Analysis} We further analyze the variation in pose estimation accuracy across datasets. RE10K’s straight, handheld trajectories allow easier relative pose estimation, while DTU’s SfM sampling trajecotry introduces more diverse viewpoint shifts and featureless background making pose estimation more challenging.

\begin{table}
    \centering
    \resizebox{\linewidth}{!}{%
        \begin{tabular}{l|cccc|cccc}
        \toprule
        \multirow{2}{*}{Method} & \multicolumn{4}{c|}{DTU-MVS (32 Frames)} & \multicolumn{4}{c}{RealEstate-10K (128 Frames)}\\
        & Latency & Speedup & AUC${}^r_{10}$ & AUC${}^t_{10}$ & Latency & Speedup & AUC${}^r_{10}$ & AUC${}^t_{10}$ \\
        \midrule
        VGGT            & 6081  & 1.00$\times$  & 0.974 & 0.981 & 90875 & 1.00$\times$  & 0.991 & 0.903 \\
        \vggts          & 1267  & 4.80$\times$  & 0.974 & 0.981 & 11407 & 7.97$\times$  & 0.990 & 0.902 \\
        ToMeSD$_{0.5}$  & 1091  & 5.57$\times$  & 0.963 & 0.971 & 7643  & 11.9$\times$  & 0.985 & 0.879 \\
        Fast VGGT       & 1998  & 3.04$\times$  & 0.950 & 0.980 & 16255 & 5.59$\times$  & 0.990 & 0.900 \\
        \rowcolor{ourcolor!30}
        \textbf{Ours}   & 788   & 7.72$\times$  & 0.958 & 0.963 & 5619  & 16.2$\times$  & 0.984 & 0.869 \\
        \midrule
        $\pi^3$         & 3197  & 1.00$\times$  & 0.965 & 0.938 & 69587 & 1.00$\times$  & 0.993 & 0.944 \\
        \rowcolor{ourcolor!30}
        \textbf{Ours}   & 824   & 3.88$\times$  & 0.946 & 0.902 & 4702  & 14.8$\times$  & 0.987 & 0.892 \\
        \midrule
        DA3-Giant       & 1573  & 1.00$\times$  & 0.963 & 0.970 & 10568 & 1.00$\times$  & 0.992 & 0.944 \\
        \rowcolor{ourcolor!30}
        \textbf{Ours}   & 709   & 2.22$\times$  & 0.937 & 0.939 & 4662  & 2.27$\times$  & 0.984 & 0.894 \\
        \midrule
        MA              & 3472 & 1.00$\times$  & 0.794 & 0.638 & 47260     & 1.00$\times$             & 0.986 & 0.868 \\
        \rowcolor{ourcolor!30}
        \textbf{Ours}   & 723 & 4.80$\times$  & 0.740 & 0.601 & 4176     & 11.32$\times$             & 0.979 & 0.839 \\
        \bottomrule
        \end{tabular}%
    }
    \vspace{-7pt}
    \captionsetup{font=footnotesize}
    \caption{Latency (\si{\milli\second}), Area Under Curve for relative rotation accuracy at 10\si{\deg} (AUC$^r_{10}$) and relative translation accuracy at 10\si{\centi\meter} (AUC$^t_{10}$) for \textbf{pose estimation} on DTU and RE10K.}
    \label{tab:pose-evaluation}
    \vspace{-8pt}
\end{table}

\subsection{Point Cloud Estimation}
\label{ssect:point-estimation}

Point cloud estimation aims to reconstruct a dense and geometrically consistent 3D representation of the scene from multi-view inputs, serving as a core task for 3D vision.

\compactpar{Metrics} We employed the completeness and accuracy from the Chamfer distance to evaluate the predicted point cloud. To remove scale and frame ambiguity, we apply a global $\mathrm{Sim}(3)$ Umeyama alignment~\citep{88573} to align the predicted point cloud to the reference point cloud. Let $P$ and $G$ denote the aligned aligned predicted and ground-truth point sets, the completeness and accuracy are defined as:
\begin{equation}
\begin{aligned}
    &\mathrm{Comp}(P,G) = \frac{1}{|G|}\sum_{g\in G}\min_{p\in P}\,\|g - p\|_2,\\
    &\mathrm{Acc}(P,G)  = \frac{1}{|P|}\sum_{p\in P}\min_{g\in G}\,\|p - g\|_2.
\end{aligned}
\end{equation}
Similar to depth evaluation, for all methods, we only evaluate regions where tokens are not merged by our method.

\compactpar{Performance} \cref{tab:point-cloud-evaluation} summarizes results on DTU and ETH3D. On DTU (32 frames), our method achieves a $7.71\times$ speedup, while maintaining essentially identical accuracy with a slight drop in completeness. 
On ETH3D (16 frames), we observe a $4.84\times$ speedup at similar performance level as VGGT.
For $\pi^3$ and MA, our method yields substantial acceleration, while the gains for DA3 are more limited due to its comparatively shallow architecture.

\compactpar{Analysis} DTU's redundant viewpoints enable efficient acceleration with minimal impact, whereas ETH3D shows \textit{improved} performance for several accelerated models, as \method removes low-confidence tokens that would otherwise introduce noise in wide-baseline reconstruction.

\begin{table}
    \centering
    \resizebox{\linewidth}{!}{%
        \begin{tabular}{l|cccc|cccc}
        \toprule
        \multirow{2}{*}{Method} & \multicolumn{4}{c|}{DTU-MVS (32 Frames)} & \multicolumn{4}{c}{ETH3D (16 Frames)}\\
        & Latency & Speedup & Comp.$\downarrow$ & Acc.$\downarrow$ & Latency & Speedup & Comp.$\downarrow$ & Acc.$\downarrow$ \\
        \midrule
        VGGT            & 6103  & 1.00$\times$ & 0.31 & 0.30 & 1752 & 1.00$\times$ & 21.0 & 20.7\\ 
        \vggts          & 1284  & 4.75$\times$ & 0.31 & 0.30 & 505  & 3.47$\times$ & 21.0 & 20.7\\
        ToMeSD$_{0.5}$  & 1095  & 5.57$\times$ & 0.32 & 0.29 & 514  & 3.41$\times$ & 19.2 & 18.7\\
        Fast VGGT       & 2001  & 3.05$\times$ & 0.33 & 0.30 & 905  & 1.94$\times$ & 20.8 & 24.5\\
        \rowcolor{ourcolor!30}
        \textbf{Ours}   & 792   & 7.71$\times$ & 0.40 & 0.31 & 362  & 4.84$\times$ & 22.3 & 25.7\\
        \midrule
        $\pi^3$         & 1577  & 1.00$\times$ & 0.45 & 0.38 & 1391 & 1.00$\times$ & 20.8 & 17.0\\
        \rowcolor{ourcolor!30}
        \textbf{Ours}   & 728   & 2.17$\times$ & 0.50 & 0.37 & 338  & 4.12$\times$ & 18.9 & 16.8\\
        \midrule
        DA3-Giant       & 1593  & 1.00$\times$ & 0.48 & 0.31 & 948  & 1.00$\times$ & 26.1 & 23.2 \\
        \rowcolor{ourcolor!30}
        \textbf{Ours}   & 687   & 2.32$\times$ & 0.53 & 0.38 & 347  & 2.73$\times$ & 27.6 & 25.0 \\
        \midrule
        MA              & 3488 & 1.00$\times$ & 0.85 & 0.50 & 1086 & 1.00$\times$ & 27.5 & 18.7\\
        \rowcolor{ourcolor!30}
        \textbf{Ours}   & 745 & 4.69$\times$ & 0.98 & 0.56 & 351 & 3.09$\times$ & 26.1 & 20.6\\
        \bottomrule
        \end{tabular}%
    }
    \vspace{-7pt}
    \captionsetup{font=footnotesize}
    \caption{Latency (\si{\milli\second}), Completeness (\si{\centi\meter}) and Accuracy (\si{\centi\meter}) for \textbf{point cloud estimation} with global $\mathrm{Sim}(3)$ alignment on DTU and ETH3D datasets. 
    On ETH3D, our method outperforms $\pi^3$ while being $4\times$ faster.
    }
    \label{tab:point-cloud-evaluation}
    \vspace{-6pt}
\end{table}

\section{Analysis}
\label{sect:analysis}

We organize our analysis around six hypotheses to examine the efficiency and effectiveness of \method.
Together, they reveal how \method achieves robust acceleration across various visual geometric models with little performance drop.

\begin{figure}
    \centering
    \includegraphics[width=\linewidth]{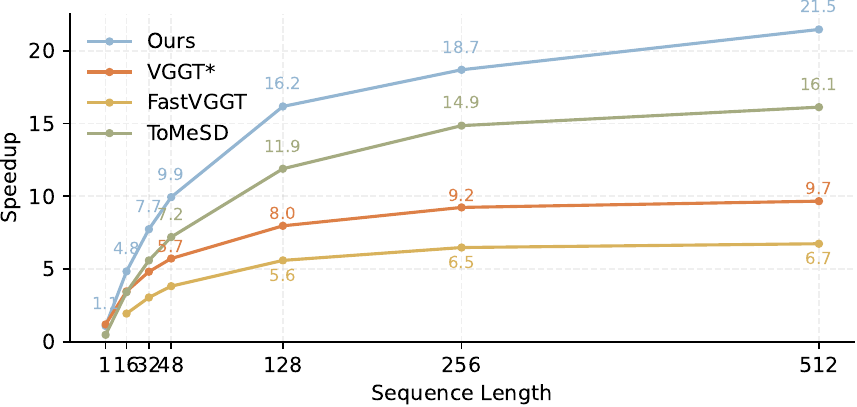}
    \vspace{-.7cm}
    \caption{
    Acceleration ratio of \method-accelerated VGGT across sequence lengths.
    The speedup increases with sequence length and reaches up to 21.5$\times$ on 512-frame input sequence.
    }
    \label{fig:speedup_seqlen}
    \vspace{-12pt}
\end{figure}

\compactpar{H1} \textbf{The speedup of \method scales with input size.}

\cref{fig:speedup_seqlen} illustrates the acceleration ratio over varying sequence lengths from $1$ to $512$, measured relative to the original VGGT baseline. 
Our method, when running on a merge ratio of $0.5$, reaches up to $21.5\times$ on 512-frame sequences, consistently outperforming FastVGGT and \vggts. 
Notably, our approach also provides measurable acceleration even on single-frame inputs, where methods such as FastVGGT offer no benefit. 
This demonstrates that the proposed acceleration mechanism is effective across different sequence lengths and scales favorably with longer inputs.

\begin{figure}
    \centering
    \includegraphics[width=\linewidth]{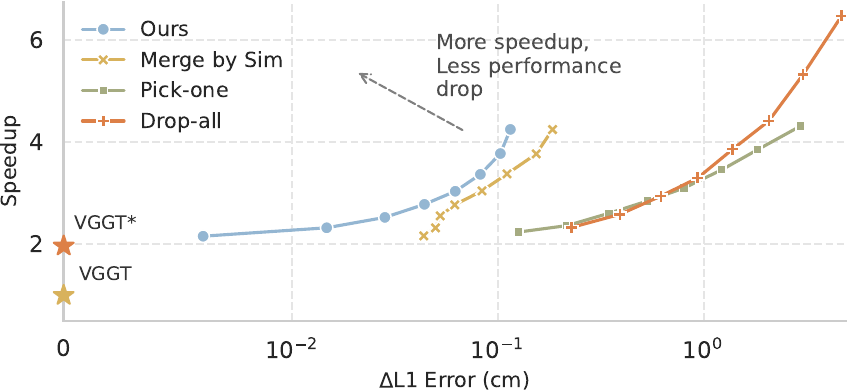}
    \vspace{-.7cm}
    \caption{Performance v. Speedup trade off curves on depth estimation by various merging thresholds on the DTU-MVS dataset. Our method delivers the optimal speed-performance tradeoff.}
    \label{fig:ablation-token-strategy}
    \vspace{-6pt}
\end{figure}

\compactpar{H2} \textbf{\method is better than similarity-based merging.}

Prior approaches, such as ToMe~\citep{bolya2023tome} and ToMeSD~\citep{Bolya_2023}, merge tokens based on cosine similarity without using confidence information. We therefore investigate how similarity based approaches perform in comparison to our method.

To this end, we establish a baseline method \textit{Merge by Sim}, which uses token similarity instead of predicted confidence for merge mask.
Specifically, tokens with average cosine similarity above $(1\!-\!p)$-percentile threshold when the merging ratio is set to $p$. 
We plot the speedup–accuracy trade-off curve on DTU multi-view depth estimation task with $p\in [0.2, 0.9]$ in \cref{fig:ablation-token-strategy}. 
Results indicate that our method yields a superior trade-off curve, achieving lower error increments than \textit{Merge by Sim} at the same speedup.

\compactpar{H3} \textbf{Merging is better than dropping or picking tokens.}

We further explore whether averaging tokens offers a stronger coalescing scheme. 
We evaluate two alternatives to merging: 1) \textit{Pick-one}, which randomly selects one token from each low-confidence group, and 2) \textit{Drop-all}, which removes low-confidence tokens entirely. 
We present the speed-accuracy trade-off curve of these setups in \cref{fig:ablation-token-strategy}. Results show that averaging via merging is significantly more robust, leading to over $10\times$ smaller performance degradation compared to the pick or drop variants.
This reveals the important role of low-confidence tokens in visual geometric transformers for providing vague contextual information.

\begin{figure}
    \centering
    \includegraphics[width=\linewidth]{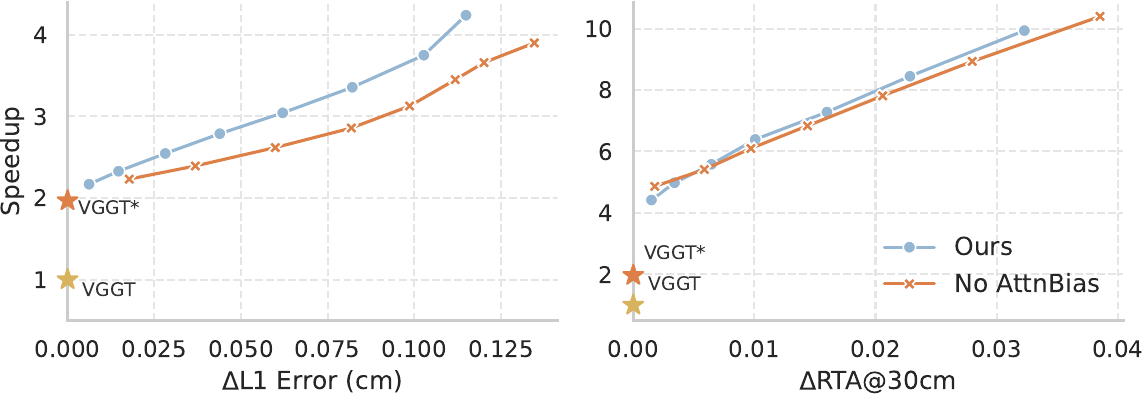}
    \vspace{-.7cm}
    \caption{
    Adding attention bias correction improves performance across tasks. On DTU multi-view depth (left), our method reduces $\Delta$L1 error by 4$\times$. On RE10K pose (right), the attention bias correction continuously offers a better performance-speed tradeoff.
    }
    \label{fig:ablation-attention-bias}
    \vspace{-6pt}
\end{figure}

\compactpar{H4} 
\textbf{Attention bias correction improves accuracy.}

To demonstrate the necessity of attention bias correction, we conducted an ablation study by removing the bias correction term from \method while keeping all other components identical. The evaluation was performed on multi-view depth estimation (DTU) and pose estimation (RE-10K). As shown in \cref{fig:ablation-attention-bias}, excluding the bias term reduces the runtime overhead by avoiding additional memory access and element-wise addition, resulting in a slight speedup. However, this comes at the cost of a substantial performance degradation. The results highlight a clear trade-off: the attention bias correction introduces minor computational overhead but significantly improves the overall accuracy.

\compactpar{H5}
\textbf{\method yields practical speedups on edge devices.}

\begin{figure}[tb]
\centering
\includegraphics[width=\linewidth]{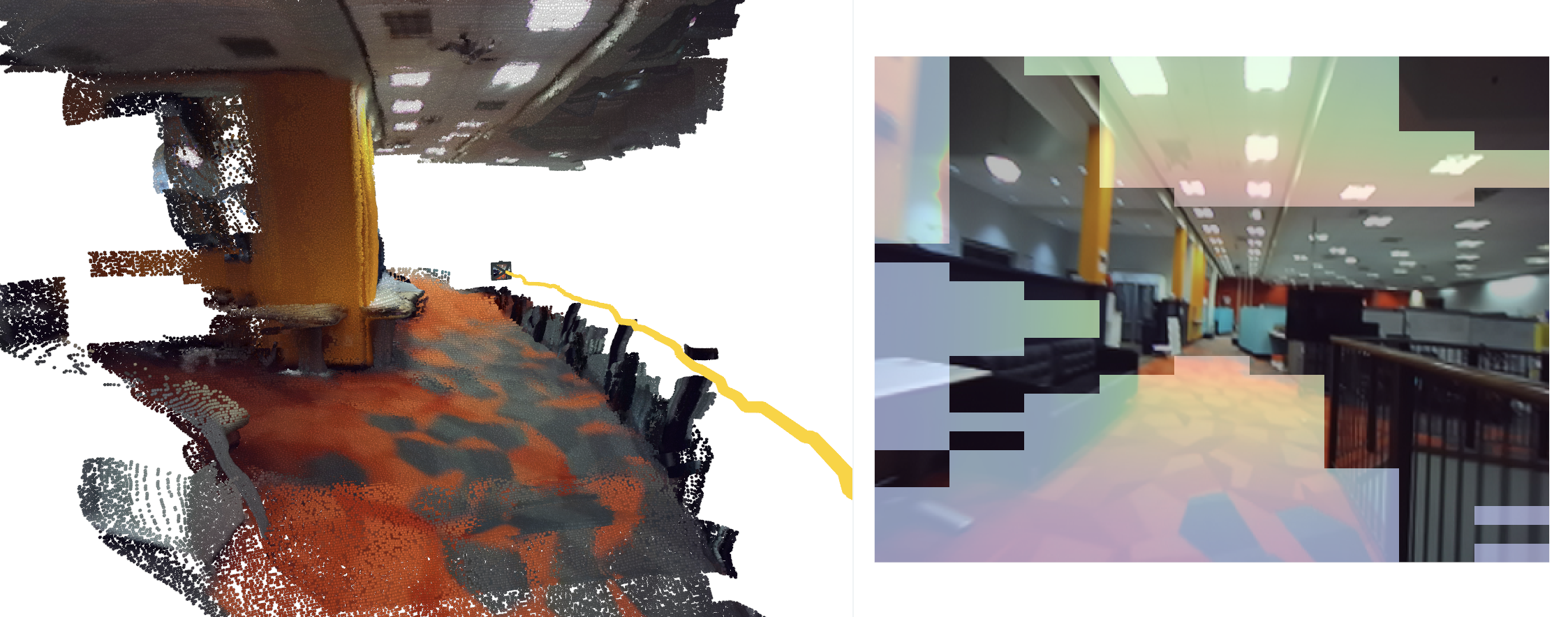}
\caption{
    Predicted 3D reconstruction (left) and depth (right) of \method-MapAnything on edge device in real time. Despite aggressive merging, the model produces consistent reconstruction. 
}
\vspace{-0.5cm}
\label{fig:hardware-demo-visualize}
\end{figure}

\begin{figure}[tb]
\centering
\setlength{\tabcolsep}{2pt}
\renewcommand{\arraystretch}{0.0}
\begin{tabular}{cc}
    \includegraphics[width=.45\linewidth]{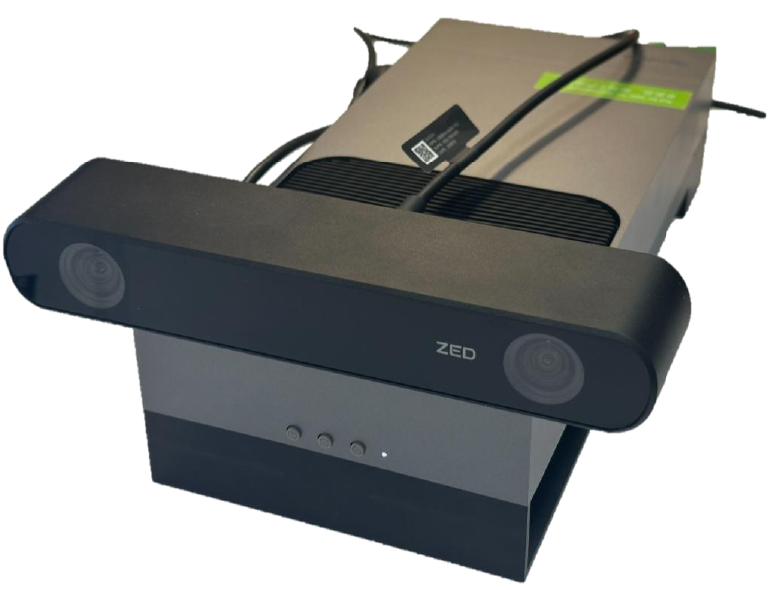} &
    \includegraphics[width=.54\linewidth]{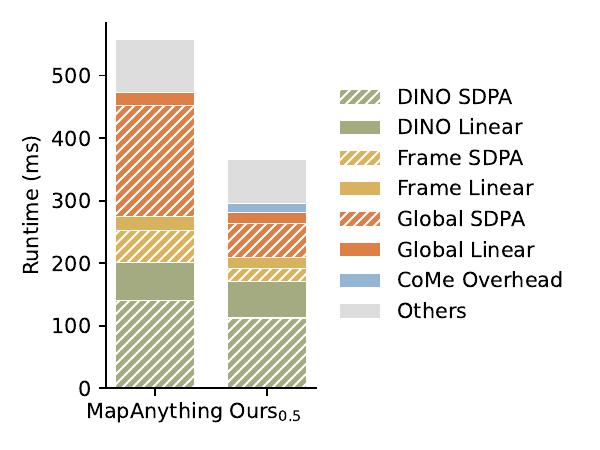}
\end{tabular}
\vspace{-0.2cm}
\caption{
    {\texttt{\small NVIDIA Jetson Thor}} and the {\texttt{\small Zed 2i}} stereo camera payload for real-world deployment test. \method accelerates the MapAnything by $1.5\times$ and achieves 3.5FPS update rate.
}
\vspace{-0.5cm}
\label{fig:hardware-demo}
\end{figure}

To assess real-world feasibility, we deploy MapAnything and our \method-accelerated variant on an NVIDIA Jetson Thor with a Zed stereo camera. The system processes streaming inputs in fixed 4-image segments and registers each reconstruction in a global frame to emulate an online visual-odometry pipeline. As illustrated in \cref{fig:hardware-demo-visualize}, the system produces consistent 3D reconstructions and accurate depth over high-confidence, unmerged regions.

Under this setup, the accelerated model achieves a 3.5 FPS update rate, providing a 1.5$\times$ speedup over the original MapAnything. This offers near real-time responsiveness under edge compute platform, demonstrating that \method delivers practical acceleration.
We show runtime decomposition and payload in \cref{fig:hardware-demo}, more detail in \cref{appendix:real-world-deployment}.

\compactpar{H6} 
\textbf{With efficient attention kernel, MLP become the new bottleneck for ViT acceleration.}

\method applies token merging not only on the scaled dot product attention (SDPA) but also the MLPs.
We hypothesize that with an efficient attention implementation, the fraction of runtime attributable to the SDPA is significantly reduced, motivating the acceleration on MLPs.

\begin{figure}
    \centering
    \includegraphics[width=\linewidth]{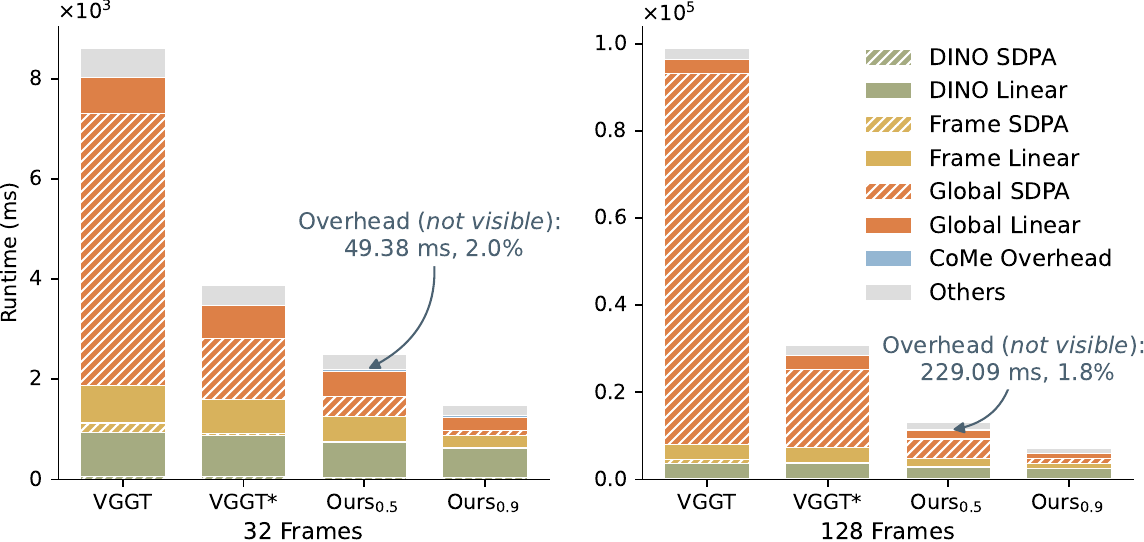}
    \vspace{-.6cm}
    \caption{
    Runtime breakdown of VGGT, \vggts, and \method accelerated \vggts with $p=\{0.5, 0.9\}$. When using efficient attention implementation, the MLPs account for a significant proportion of the runtime in network inference. 
    \method can accelerate all modules in the network with minimal system overhead.
    }
    \label{fig:runtime-analysis}
    \vspace{-16pt}
\end{figure}

In \cref{fig:runtime-analysis}, we show the detailed runtime decomposition of  VGGT, \vggts, and \method with merge ratio $p=\{0.5, 0.9\}$ and sequence lengths of 32 and 128. In vanilla VGGT, SDPA dominates the runtime. However, with the efficient SDPA, linear layers still take a considerable portion of inference time even under long, 128-frame input.
This trend highlights 
MLP become the new bottlenecks as the SDPA is thoroughly optimized by previous works. 

Moreover, the efficient implementation in \cref{ssect:efficient-impl} ensures \method incurs negligible overhead in \cref{fig:runtime-analysis}, accounting for $\approx2\%$ of inference time on the accelerated VGGT.

\section{Conclusion}
\label{sect:conclusion}

We presented Confidence-Guided Token Merging (\method), a novel method to accelerate visual geometric transformers by merging low-confidence tokens guided by a distilled confidence predictor without retraining or finetuning the base model. When applied on VGGT, \method achieves up to 21.5$\times$ speedup while preserving accuracy in depth, pose, and point estimation. We also demonstrate that \method is generalizable by applying it to four state-of-the-art models, where \method demonstrates exceptional acceleration while maintains comparable performance with original models. 

Future extensions of this work include generalize the merging guidance signal from geometric-confidence to task-relevance for vision-language and vision-language-action models, applying merging to time dimension for streaming inputs in robotics application, and utilizing \method in the training time to improve training efficiency.

{
    \clearpage\small
    \bibliographystyle{ieeenat_fullname}
    \bibliography{main}
}

\iftoggle{cvprfinal}{
\clearpage
\appendix
\twocolumn[
\begin{center}
    \vspace*{1.0em}
    {\Large Supplementary Materials for\\[0.5em]
    \textbf{\method: Confidence-Guided Token Merging for Visual Geometric Transformers}}
    \vspace{1.25em}
\end{center}
]
\begin{figure}[tb]
\centering
\setlength{\tabcolsep}{2pt}
\renewcommand{\arraystretch}{0.0}
\begin{tabular}{cc}
VGGT & Co-Me + VGGT \\[4pt]
\includegraphics[width=0.48\linewidth]{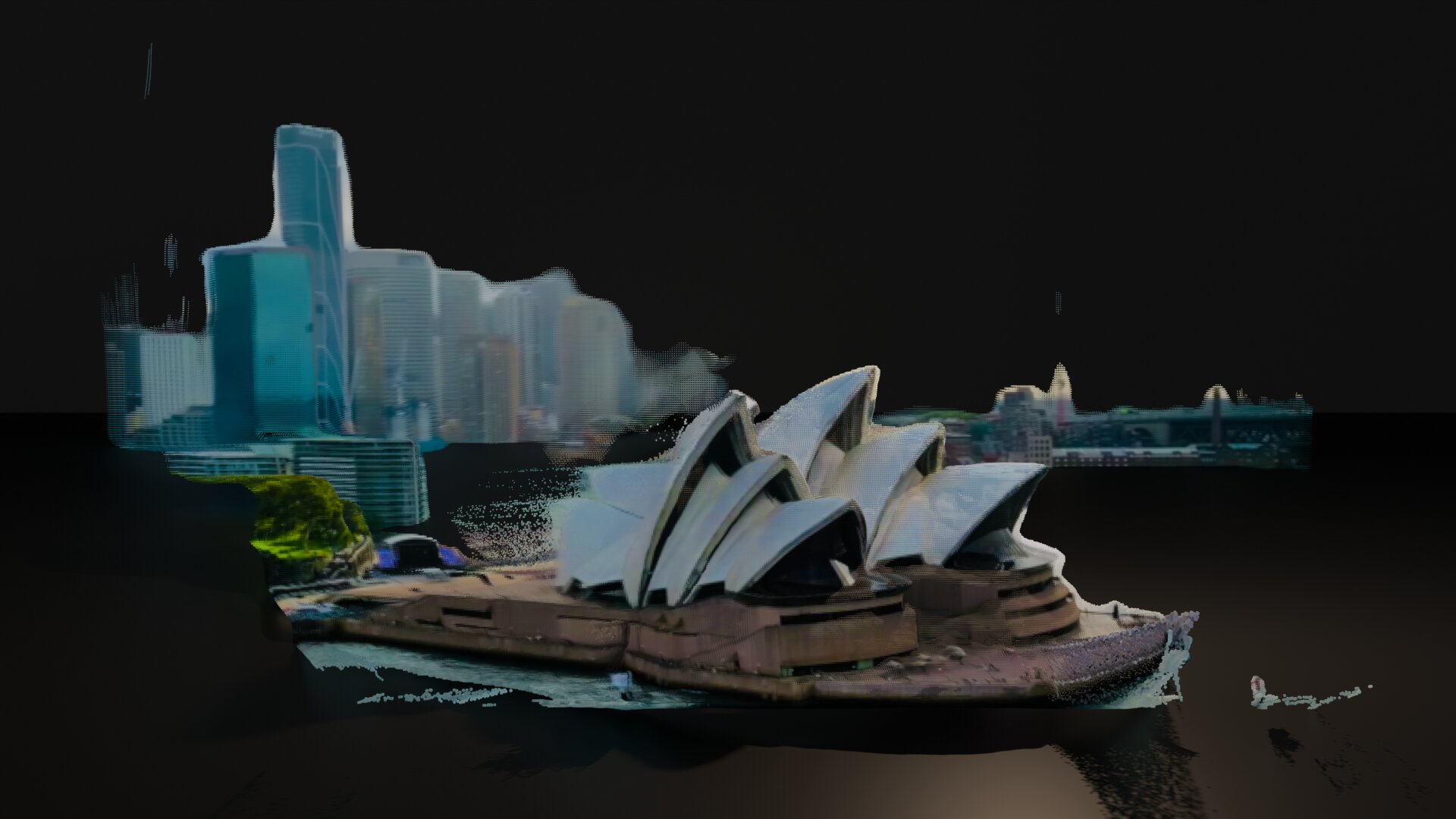} &
\includegraphics[width=0.48\linewidth]{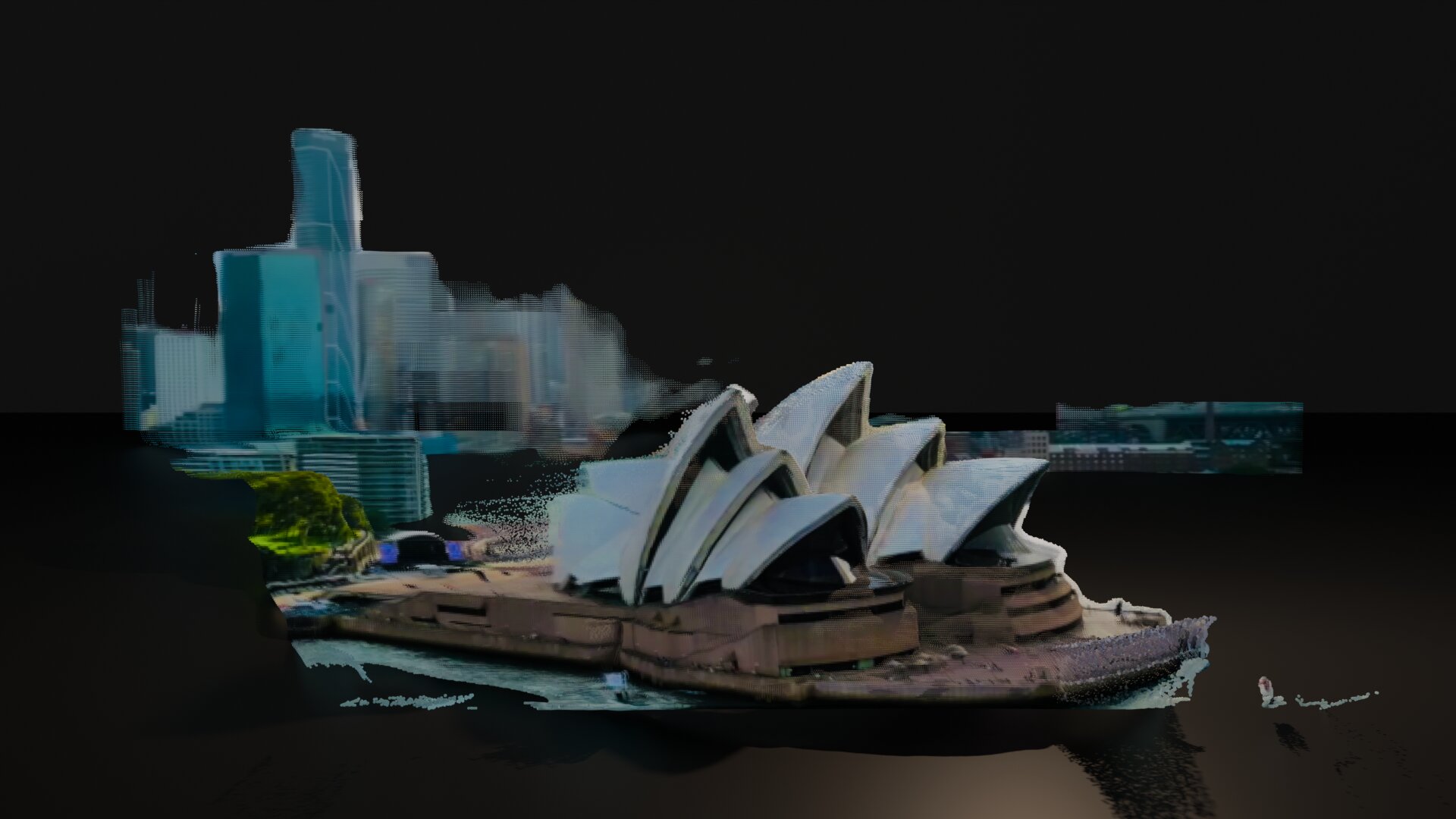} \\[4pt]
\includegraphics[width=0.48\linewidth]{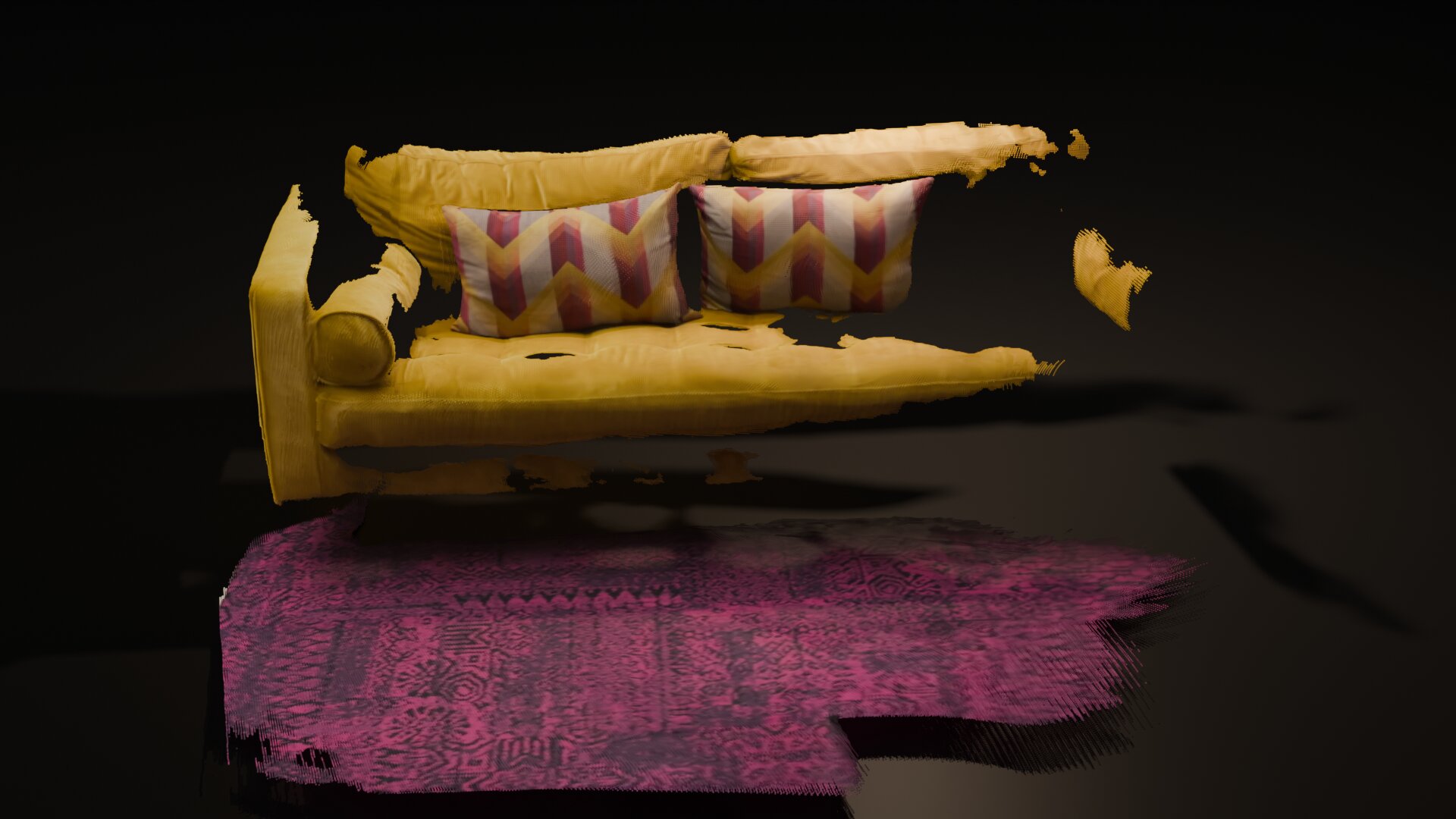} &
\includegraphics[width=0.48\linewidth]{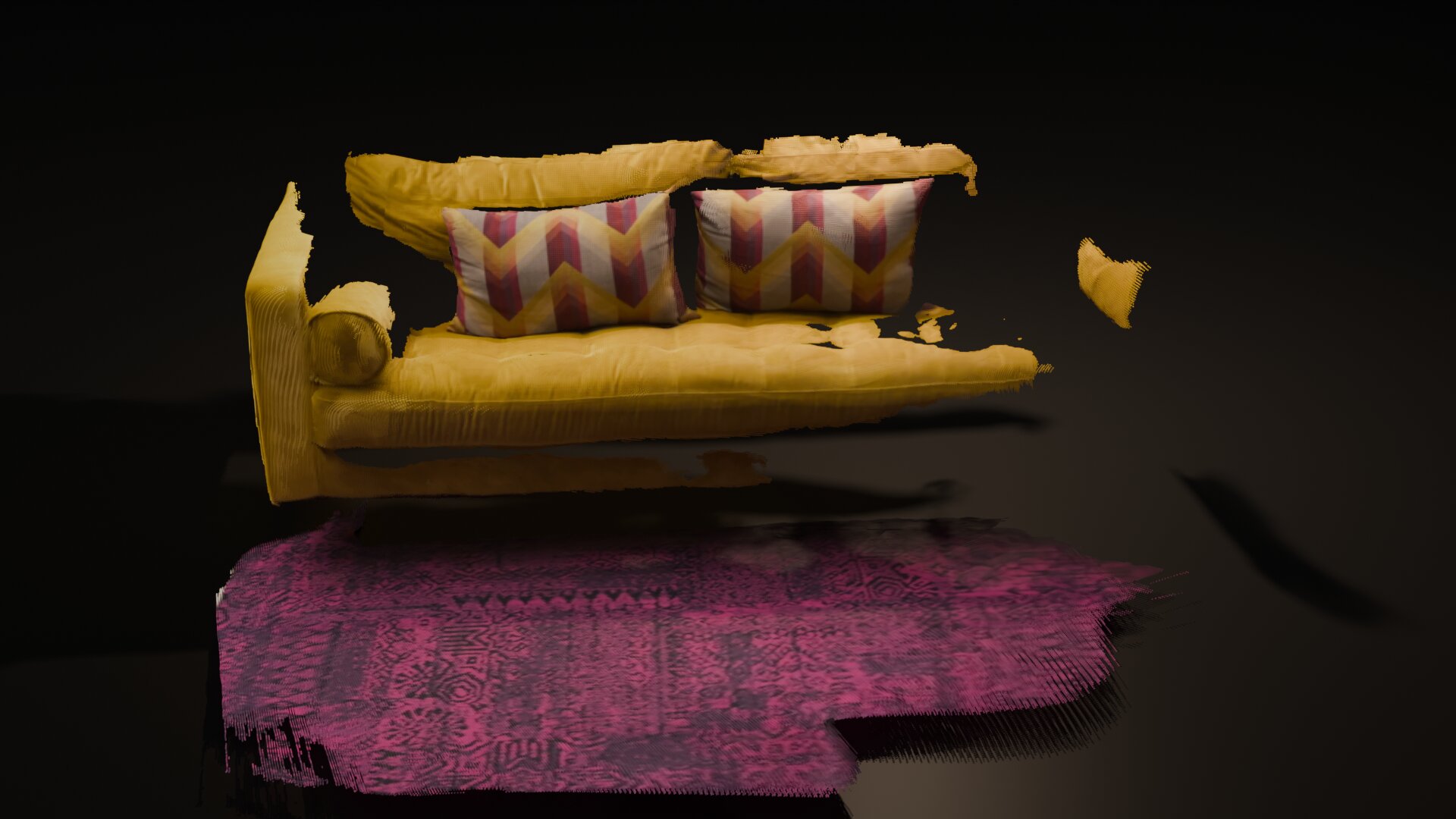} \\[4pt]
\includegraphics[width=0.48\linewidth]{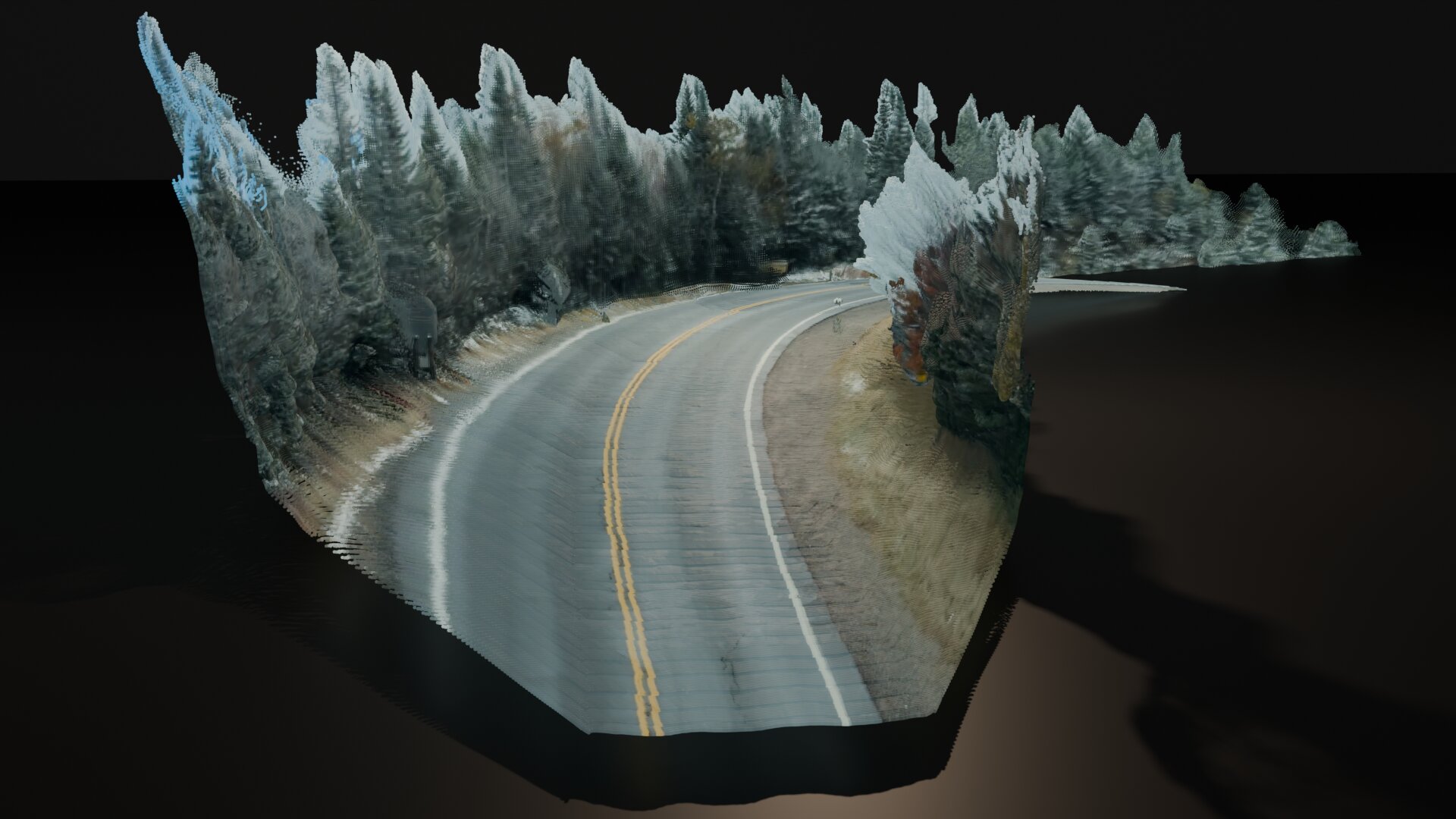} &
\includegraphics[width=0.48\linewidth]{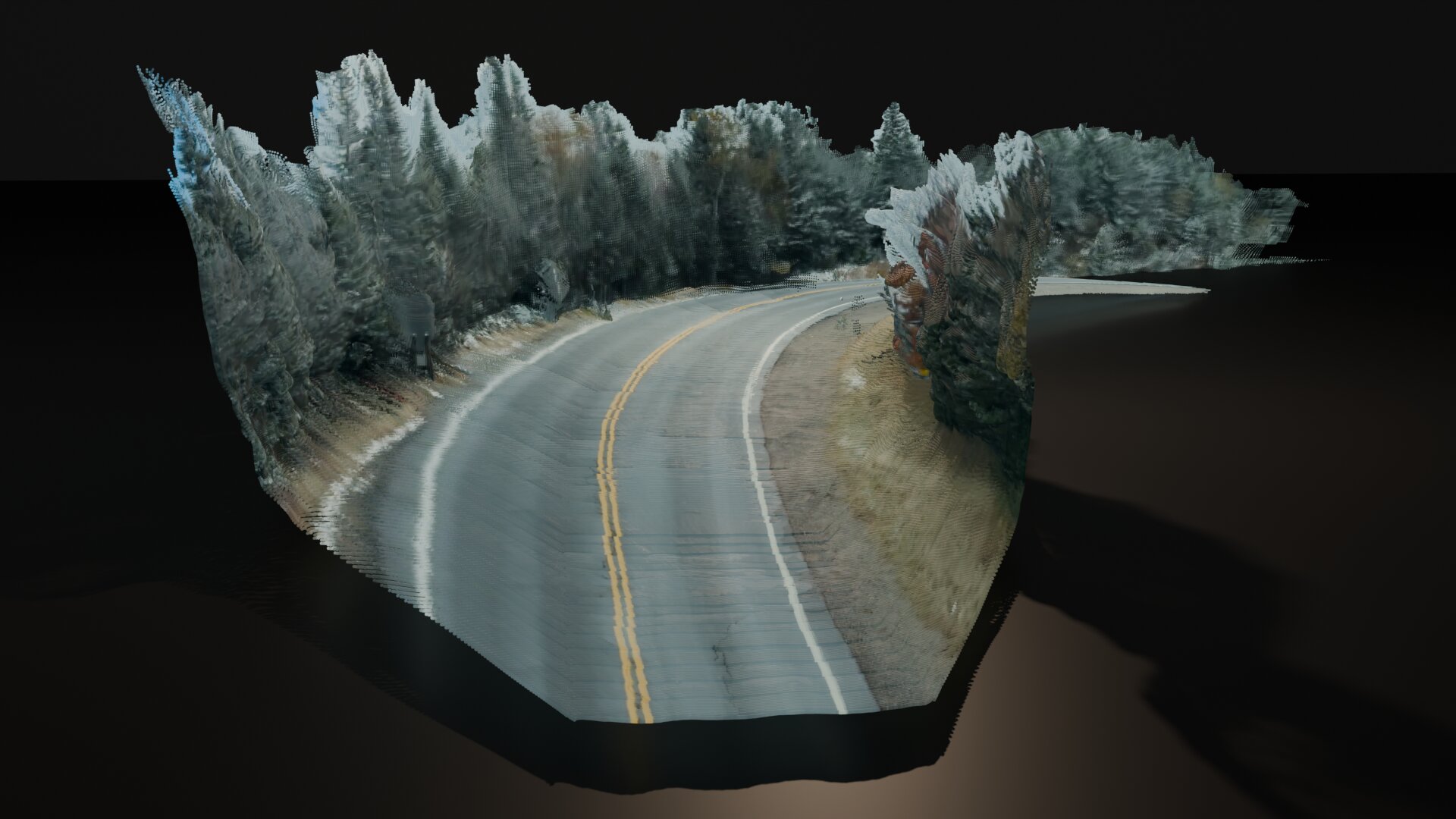} \\[4pt]
\includegraphics[width=0.48\linewidth]{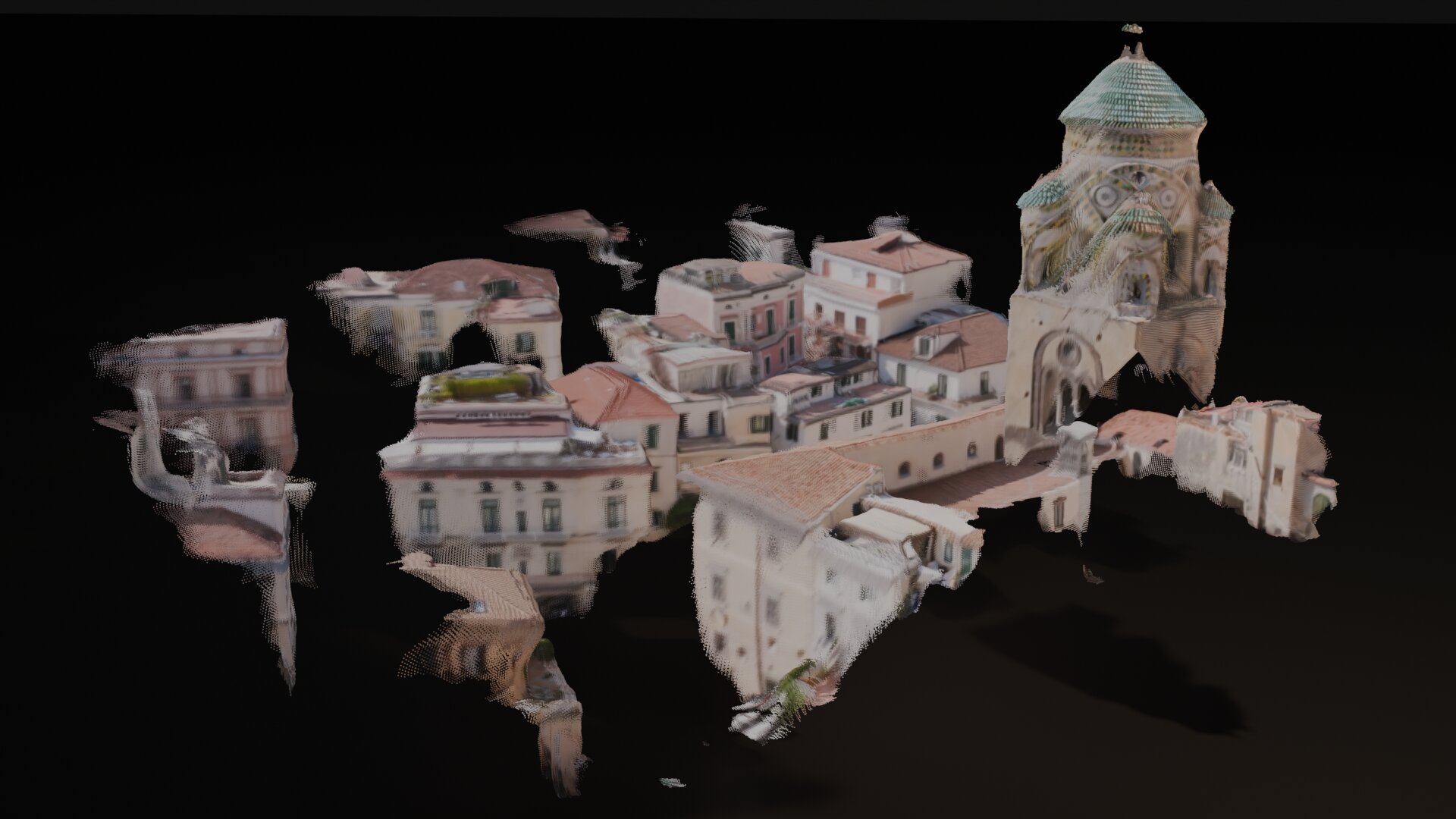} &
\includegraphics[width=0.48\linewidth]{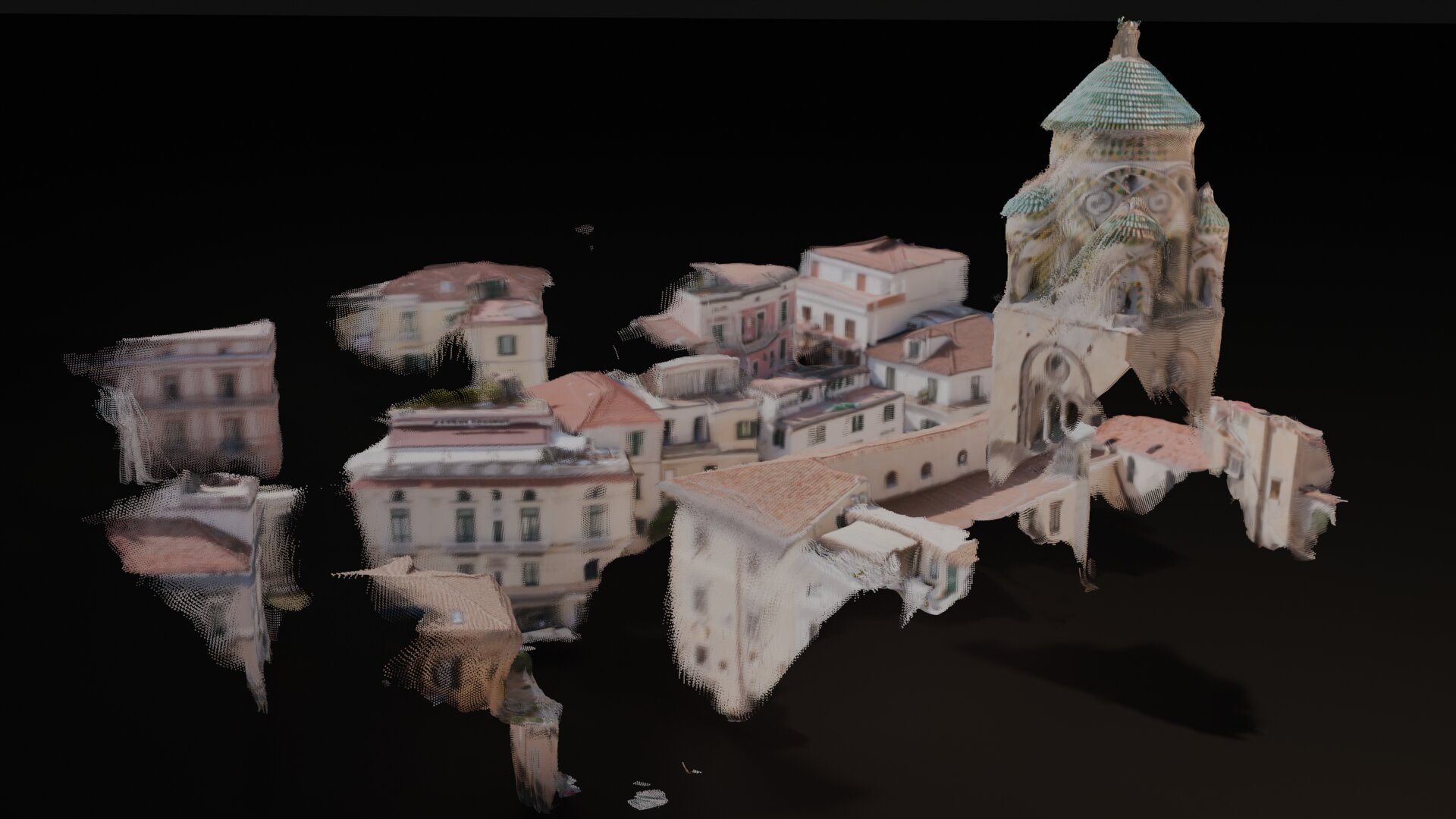} \\[4pt]
\end{tabular}
\caption{
Qualitative comparison between VGGT (left) and \method-accelerated VGGT (right). Best viewed digitally.
}
\label{fig:compare_grid}
\end{figure}

\begin{figure}[tb]
\centering
\setlength{\tabcolsep}{2pt}
\renewcommand{\arraystretch}{0.0}
\begin{tabular}{cc}
MapAnything & Co-Me + MapAnything \\[4pt]
\includegraphics[width=0.48\linewidth]{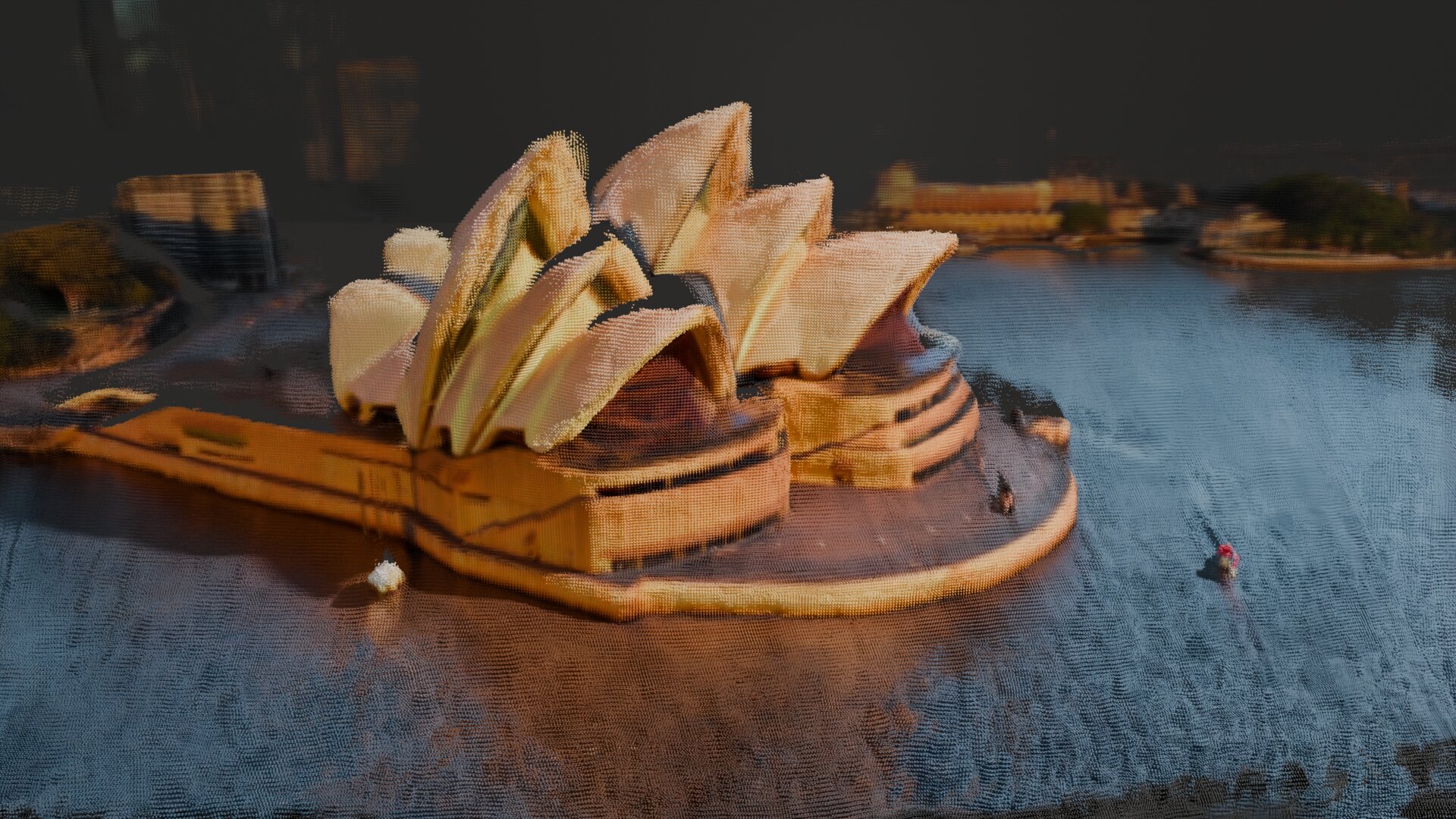} &
\includegraphics[width=0.48\linewidth]{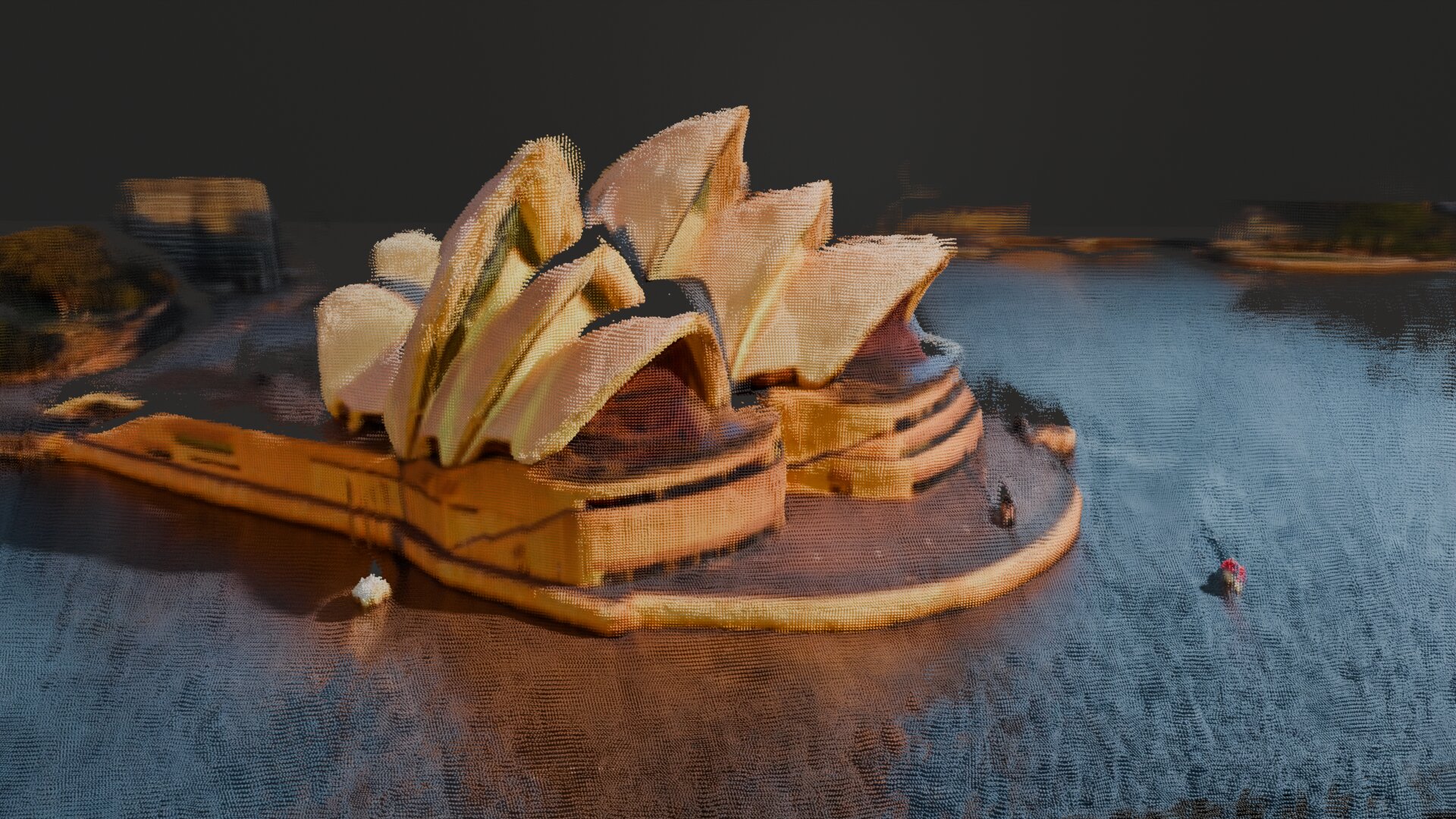} \\[4pt]
\includegraphics[width=0.48\linewidth]{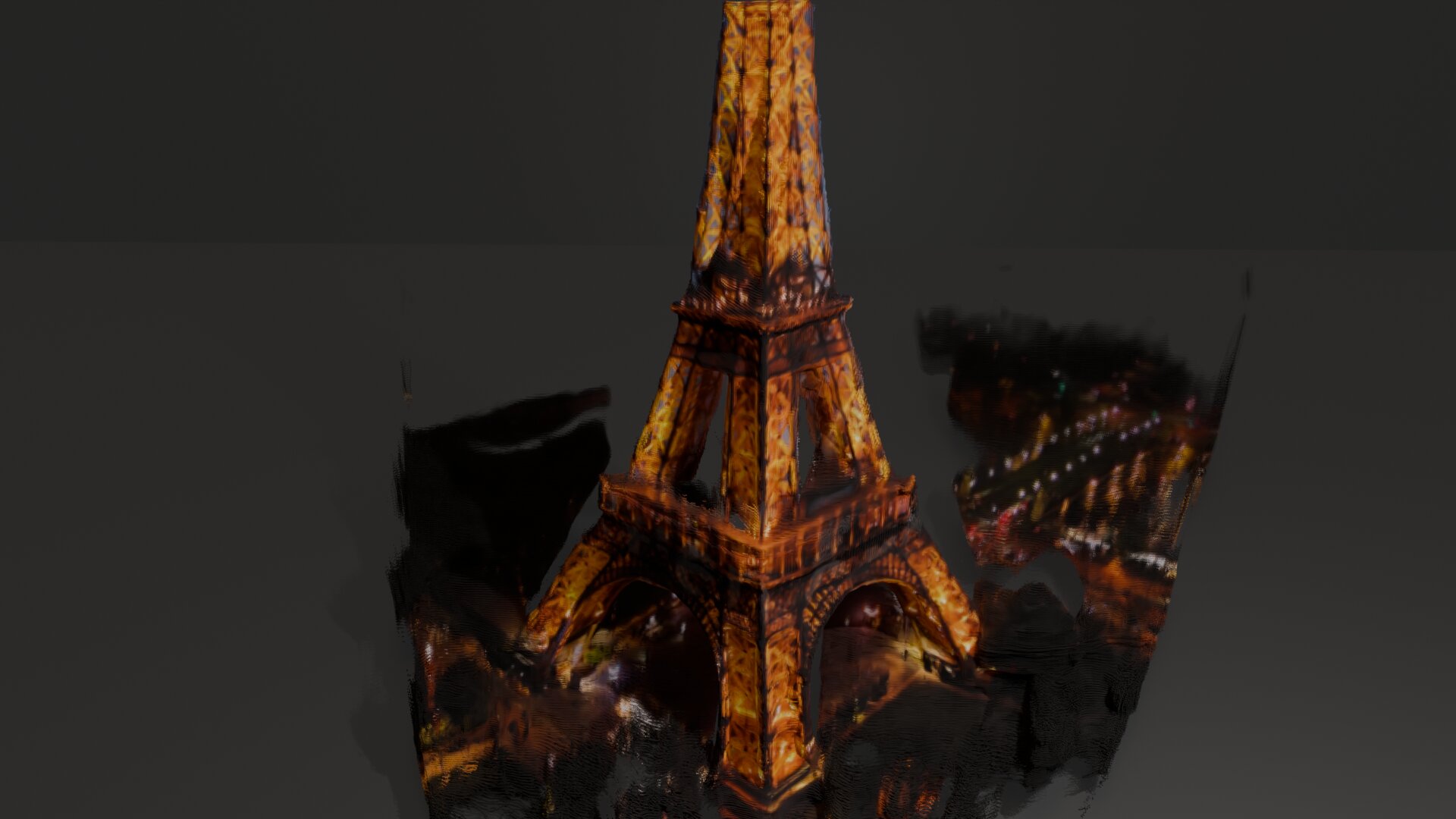} &
\includegraphics[width=0.48\linewidth]{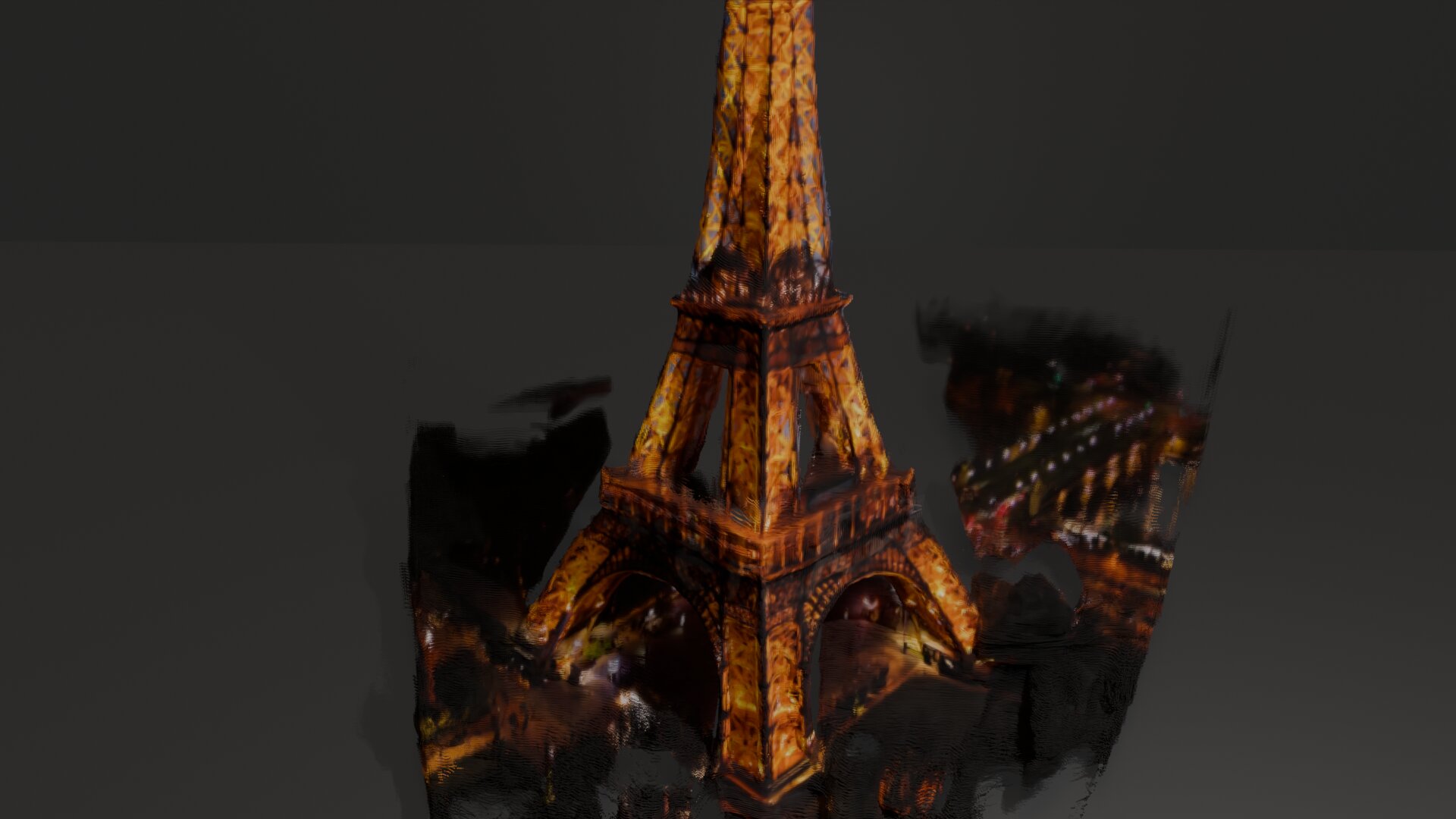} \\[4pt]
\includegraphics[width=0.48\linewidth]{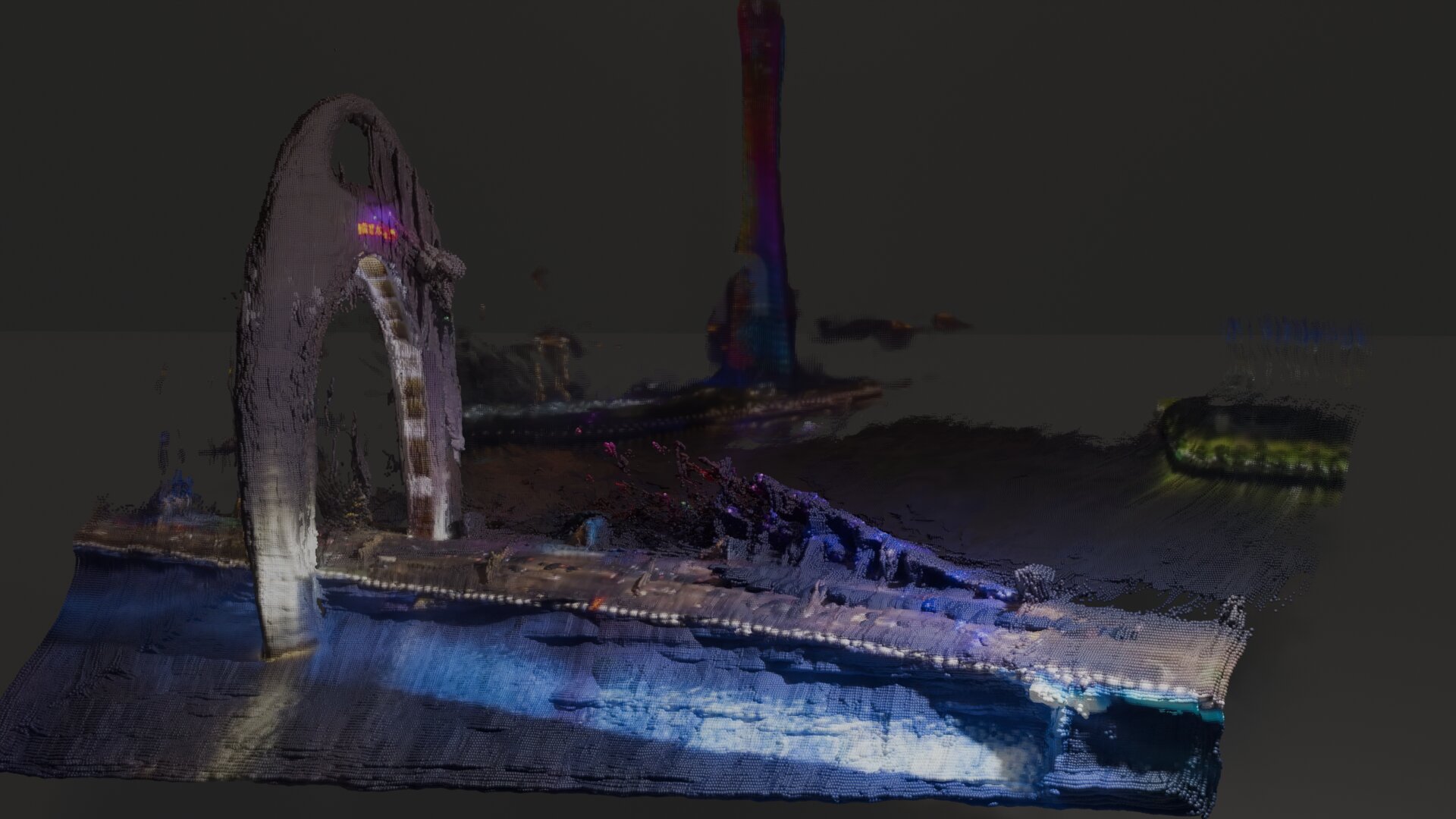} &
\includegraphics[width=0.48\linewidth]{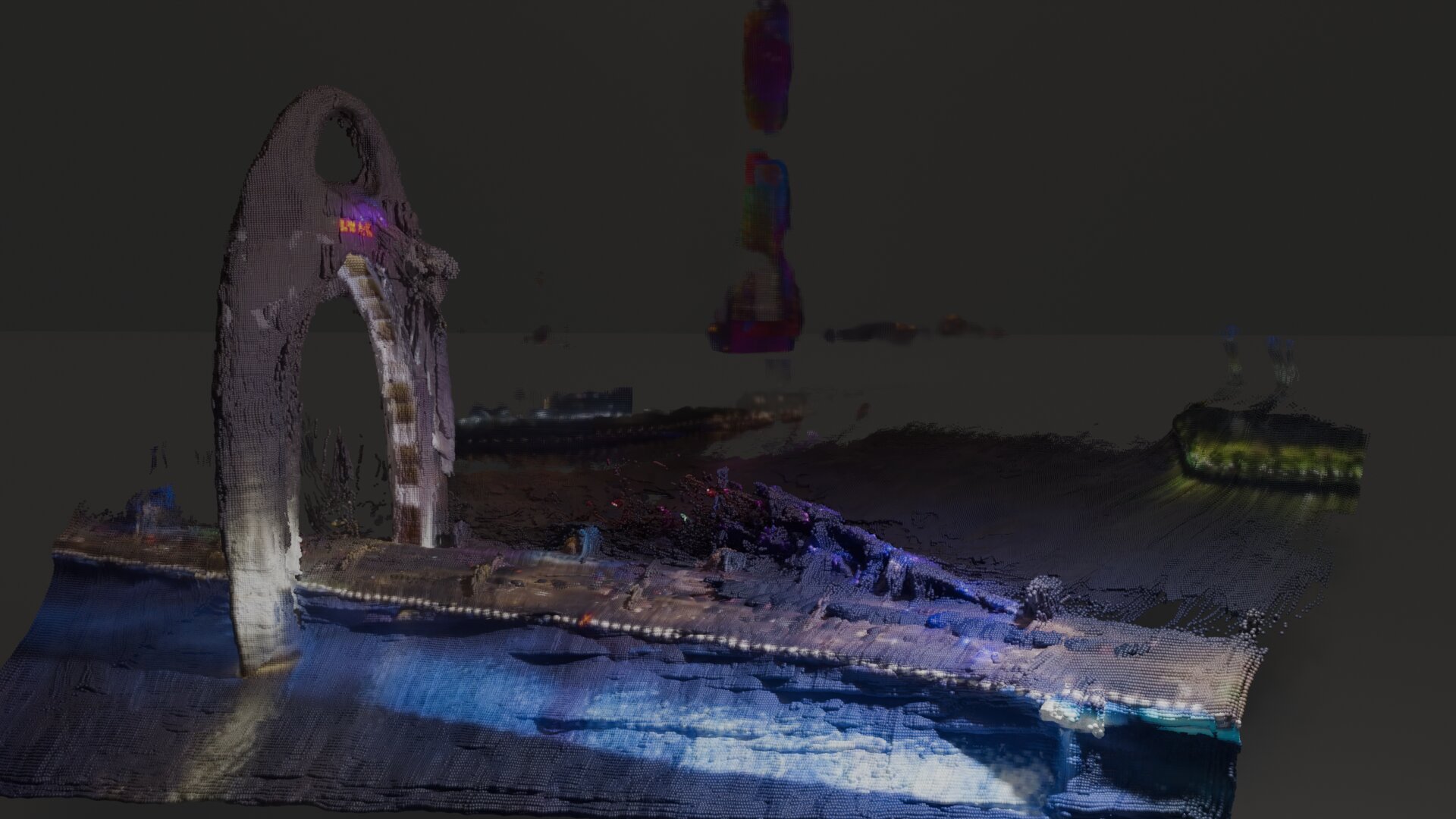} \\[4pt]
\includegraphics[width=0.48\linewidth]{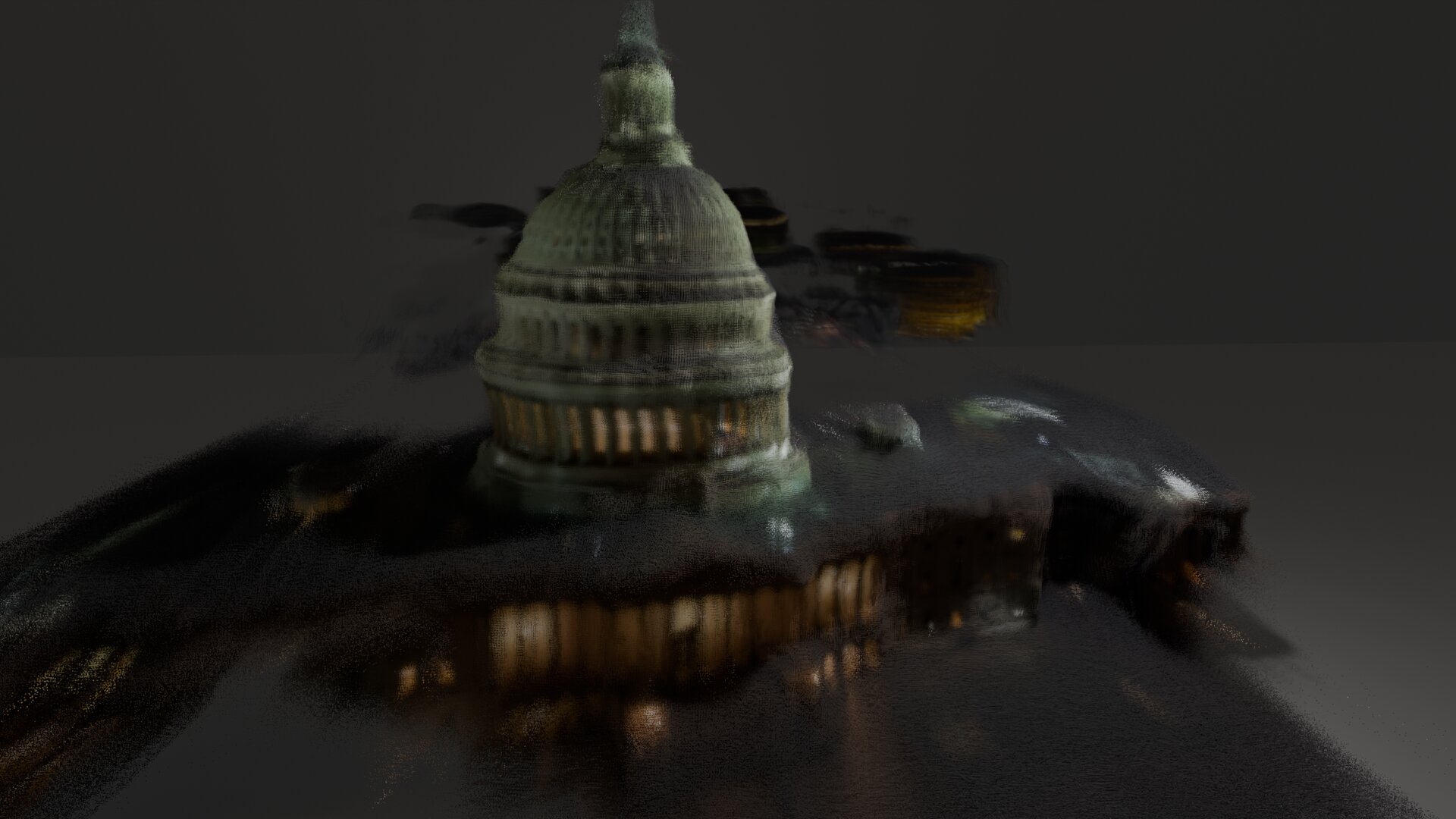} &
\includegraphics[width=0.48\linewidth]{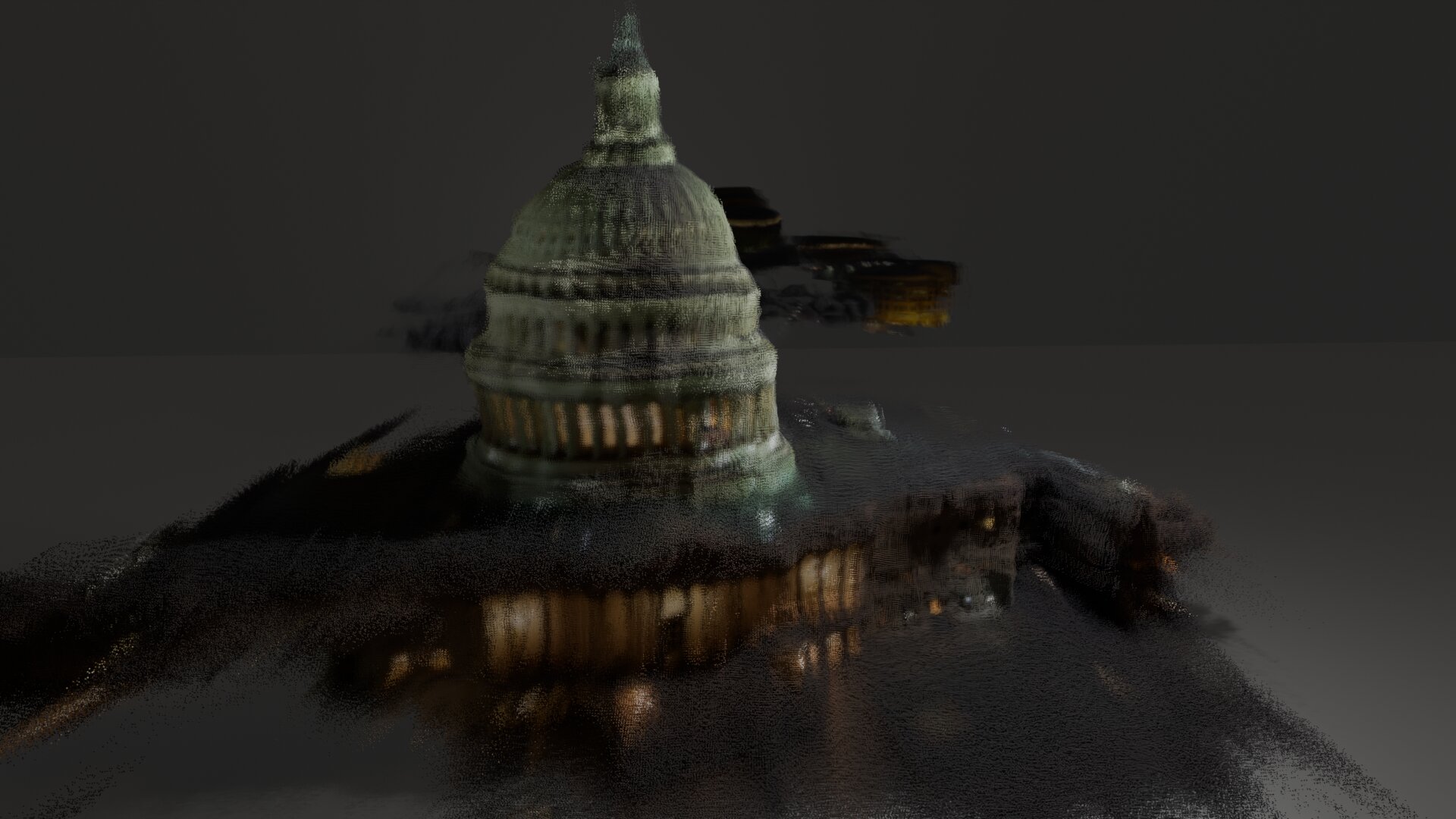}
\end{tabular}
\caption{
Qualitative comparison between MapAnything (left) and \method-accelerated version (right). Best viewed digitally.
}
\label{fig:compare_grid_mapanything}
\end{figure}

\section{Qualitative Results}

In this section we present qualitative results of \method-accelerated VGGT and MapAnything. Specifically all results are created with the exact same configuration in \cref{sect:experiments} without finetuning or modification.

\subsection{Success Cases}

\compactpar{VGGT} In \cref{fig:compare_grid}, we show a qualitative comparison between VGGT (left) and \method-accelerated VGGT (right) across eight representative scenes. \method preserves the global scene structure and fine-grained geometry, including planar surfaces and prominent edges, despite operating with significantly fewer tokens.
Minor differences appear primarily along the boundaries between high-confidence foreground regions and low-confidence background areas.
These examples illustrate that confidence-guided merging maintains reconstruction fidelity with reduced computation.

\compactpar{MapAnything}
\Cref{fig:compare_grid_mapanything} reports qualitative reconstructions from MapAnything and its \method-accelerated variant across four diverse outdoor scenes. Despite aggressive token reduction, the accelerated model retains the characteristic large-scale structure that MapAnything recovers—such as façade geometry, smooth water surfaces, and distant skyline contours.
Most observable differences are confined to peripheral regions where texture cues are weak or depth ambiguity is intrinsic to the input views. In these areas, \method may slightly simplify fine-scale geometry, but the dominant scene layout and salient landmarks remain stable.
These results show that token merging integrates cleanly with the MapAnything pipeline, preserving its strong global consistency while reducing inference cost.

\subsection{Failure Modes}

\begin{figure}[tb]
\centering
\setlength{\tabcolsep}{2pt}
\renewcommand{\arraystretch}{0.0}
\begin{tabular}{cc}
VGGT & Co-Me + VGGT \\[4pt]
\includegraphics[width=0.48\linewidth]{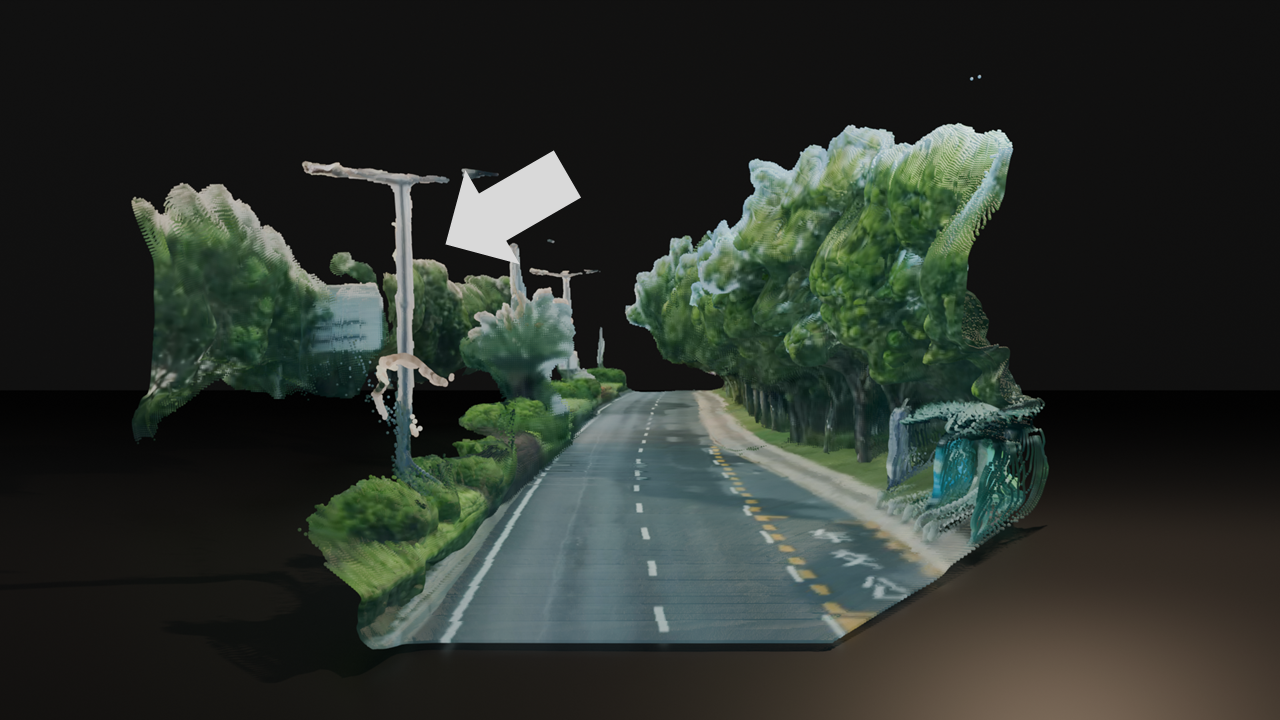} &
\includegraphics[width=0.48\linewidth]{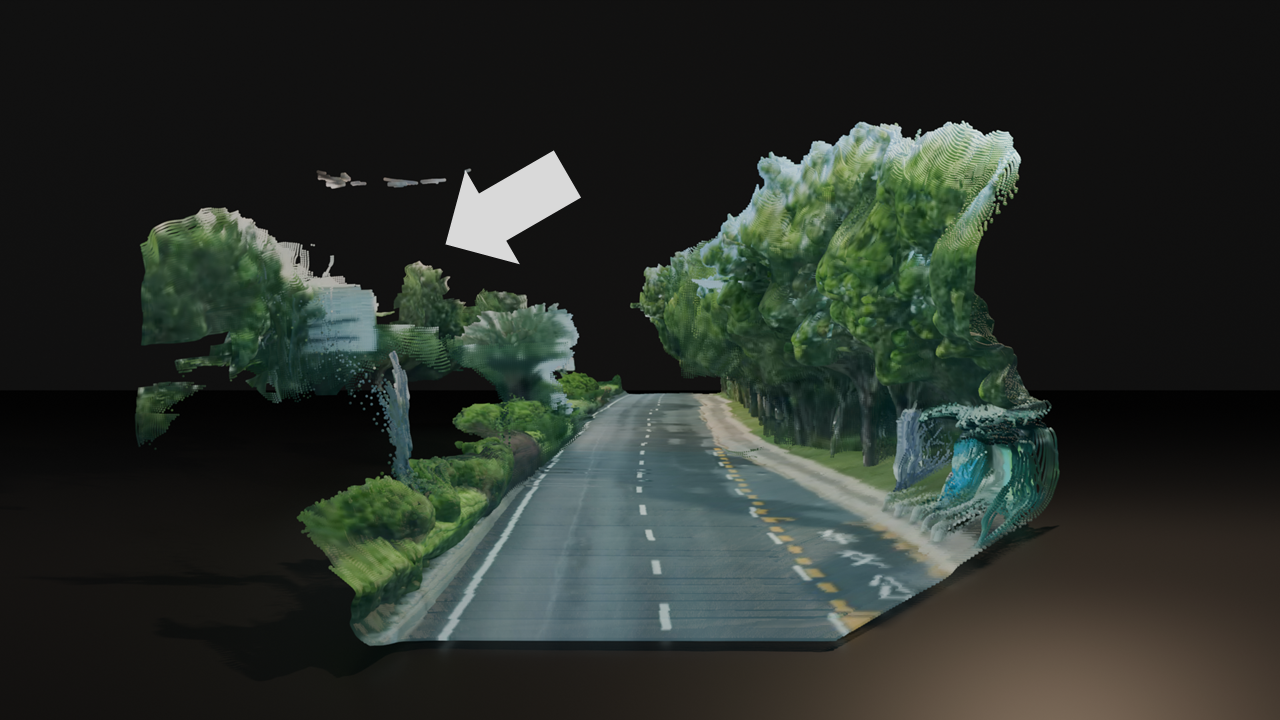} \\[4pt]
\includegraphics[width=0.48\linewidth]{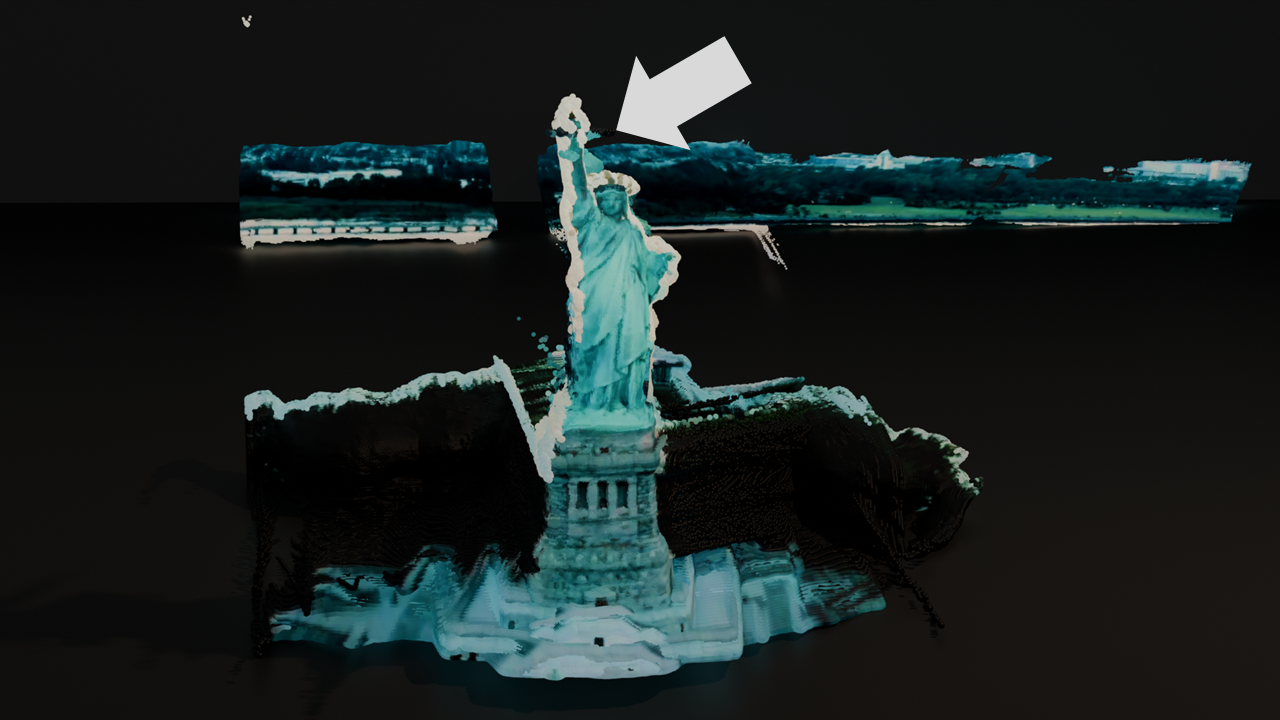} &
\includegraphics[width=0.48\linewidth]{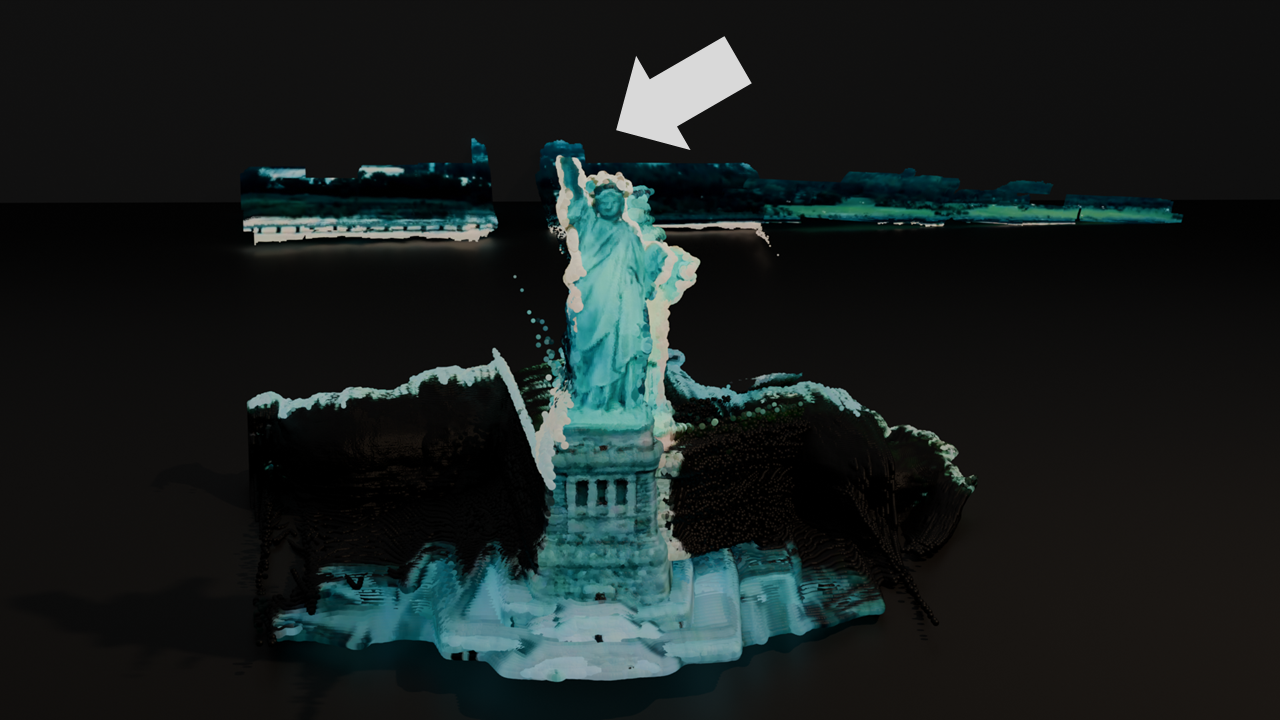} \\[4pt]
\end{tabular}
\caption{
Failure cases of \method-accelerated VGGT. Comparison between VGGT (left) and \method-accelerated VGGT (right). 
Arrows indicate corrupted thin structures after token merging.
}
\label{fig:compare_grid_failure}
\end{figure}

\Cref{fig:compare_grid_failure} highlights scenarios where \method introduces noticeable degradation. In both examples, the lost geometry corresponds to thin, high-frequency structures that occupy a small portion of the corresponding token. When these regions have low predicted confidence, merging discards their local resolution enough that the downstream decoder oversmooths the structure, causing incomplete reconstruction of the streetlight pole and the Statue of Liberty’s raised arm. While these elements do not affect the global scene layout, they reveal a limitation of confidence-guided merging in handling small or elongated objects.

\section{Additional Experiments}
\label{appendix:additional_analysis}

\subsection{Confidence Distillation Layer Ablation}
\label{appendix:additional_analysis_insert_layer}

To investigate where the confidence predictor should be inserted within the ViT backbone, we trained the predictor on features extracted from different encoder layers of VGGT under identical training setups. 
\cref{fig:layer-loss} illustrates the ranking loss curves for predictors attached to layers 6, 9, 12, 15, 18, and 21 respectively. 
We observe that the predictor distilled from layer 15 achieves the lowest ranking loss across all layers. 
Earlier layers (e.g., 6, 9) provide insufficient semantic and geometric cues, leading to noisy confidence estimates, while later layers (e.g., 18, 21) have stronger geometric reasoning but reduced token diversity, which limits generalization and causes slower convergence. 
Therefore, we use the layer-15 configuration in all experiments, as it provides an optimal trade-off between confidence ranking accuracy and computational overhead.

\begin{figure}
    \centering
    \includegraphics[width=\linewidth]{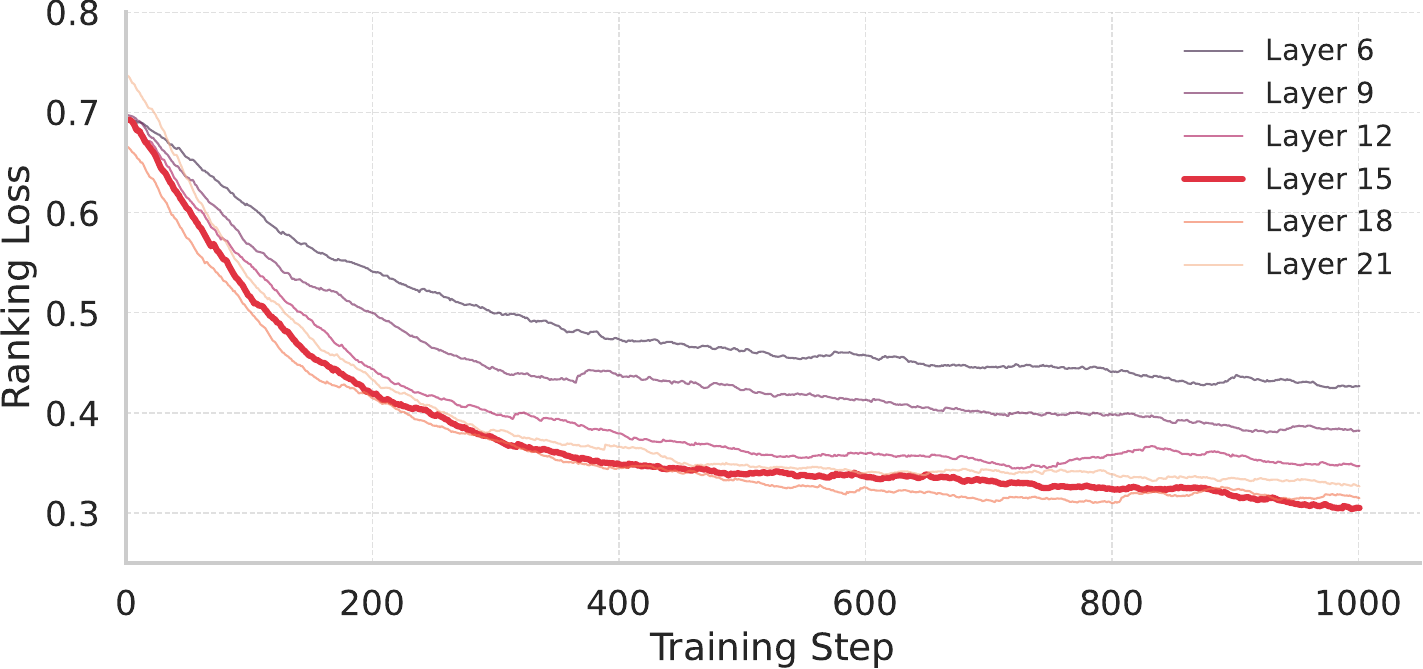}
    \caption{
        Distillation loss of confidence predictors distilled from various VGGT encoder layers.
        Layer 15 yields the lowest loss, indicating that mid-level encoder features contain the most information for confidence estimation.
        For readability, curves are smoothed with an exponential moving average with factor of 0.99.
    }
    \label{fig:layer-loss}
\end{figure}

\subsection{Confidence Distillation Loss Ablation}
\label{appendix:confidence-distill-loss-ablate}
\begin{figure}
    \centering
    \includegraphics[width=\linewidth]{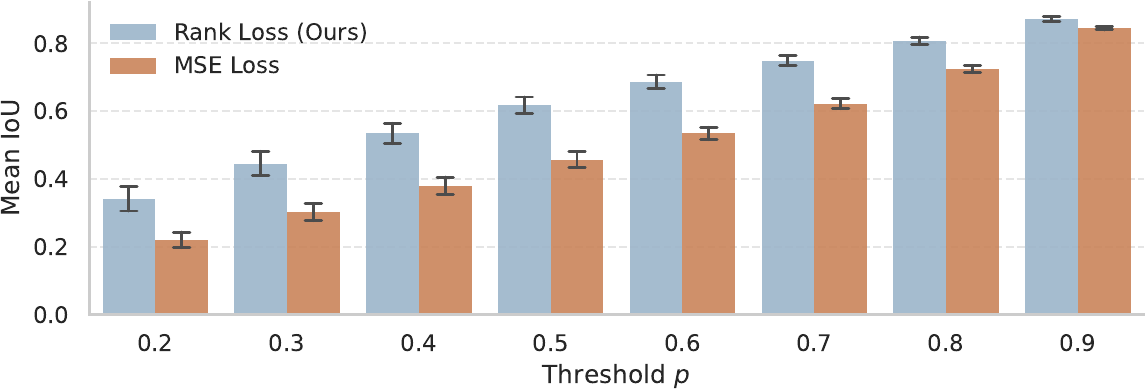}
    \caption{Confidence distillation with ranking loss achieves significantly higher IoU on DTU dataset than MSE loss. Error bar shows the 95\% confidence interval. 
    }
    \label{fig:ablation-pairwise-loss}
\end{figure}

In the \cref{ssect:confidence-distill}, we replace the MSE objective with a ranking loss 
that supervises the relative ordering of token confidences. 
To validate the effectiveness of loss formulation, we conduct an ablation by retraining the model using MSE under identical settings. We then compare the resulting predictions by measuring the intersection-over-union (IoU) between the top-$p$ merge masks derived from the distilled predictor and those obtained from the full VGGT model, with $p \in [0.2, 0.9]$. The IoU metric is defined as:
\begin{equation}
    \mathrm{IoU} = \frac{|\mathcal{M}_{\text{pred}} \cap \mathcal{M}_{\text{gt}}|}{|\mathcal{M}_{\text{pred}} \cup \mathcal{M}_{\text{gt}}|},
\end{equation}
where $\mathcal{M}_{\text{pred}}, \mathcal{M}_{\text{gt}}$ are the predicted and reference masks.

In \cref{fig:ablation-pairwise-loss}, we can see that the ranking loss consistently outperforms MSE, 
demonstrating that supervising the relative ordering of confidences is more effective than regressing the confidence numerically for predicting merge masks.

\subsection{Token Group Size Ablation}

\begin{figure}
    \centering
    \includegraphics[width=\linewidth]{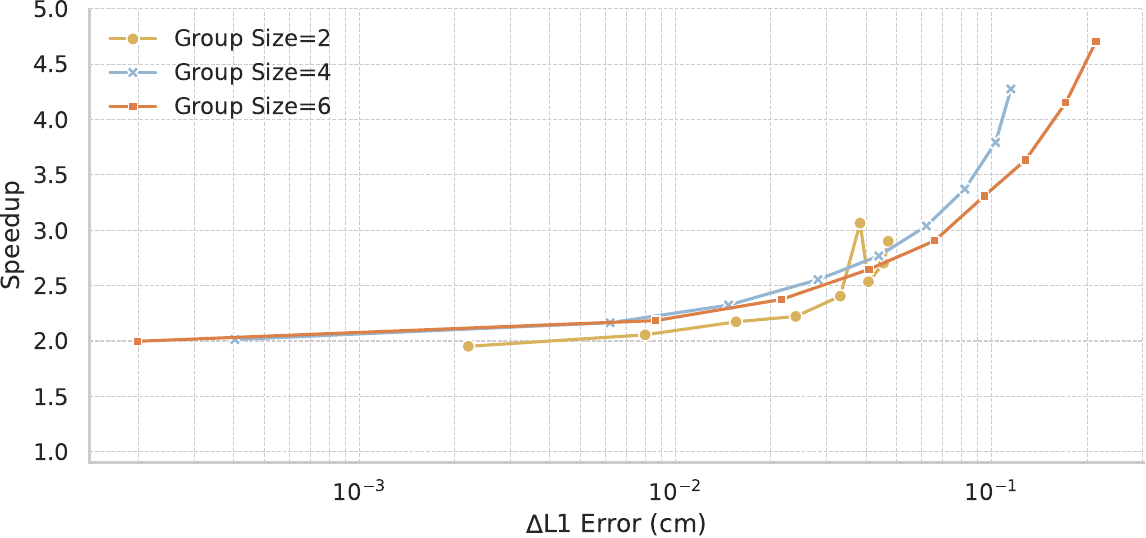}
    \caption{
        Speedup-accuracy trade-off of \method-accelerated VGGT across various token group sizes on multi-view depth estimation (DTU-MVS, 32 frames). Smaller group sizes yield slightly better accuracy, while larger groups provide higher acceleration. Curves are plotted on a log-scaled error axis for clarity.
    }
    \label{fig:grp-size-ablate}
\end{figure}

To evaluate the influence of token group size on the speed–accuracy trade-off, we tested Co-Me with group sizes of 2, 4, and 6 under identical merging ratios on DTU-MVS (32 frames). As shown in Fig. \cref{fig:grp-size-ablate}, smaller group sizes generally offer better accuracy retention for a given speedup, as they introduce finer-grained control and less information loss over which merged tokens. In contrast, larger groups provide stronger acceleration due to more aggressive token reduction, but incur slightly higher reconstruction error. Overall, group size 4 achieves the best balance between efficiency and accuracy and is therefore used for all experiments in \cref{sect:experiments,sect:analysis}.

\section{Edge Compute Deployment}
\label{appendix:real-world-deployment}

In \cref{fig:hardware-demo}, we illustrate the real-world deployment setup and runtime profile used to evaluate edge performance. An NVIDIA Jetson Thor runs MapAnything and our \method-accelerated variant while receiving stereo input from a Zed 2i camera. The system groups incoming frames into fixed segments of four images and accumulates the resulting reconstructions in a global world coordinate frame, effectively simulating a streaming visual-odometry pipeline.

The stacked runtime bars in \cref{fig:hardware-demo} decompose per-segment latency into DINO, frame-level, and global attention components, linear projections, \method overhead, and other operations. Applying \method shrinks the attention-dominated portions while adding only a small confidence-prediction cost, yielding an overall 1.5$\times$ reduction in end-to-end runtime. On this platform, processing 4-image segments reaches 3.5 FPS, providing near real-time responsiveness under edge-compute constraints while preserving the stable 3D geometry observed in \cref{sect:analysis}, H5.

}{}
\end{document}